\documentclass[letterpaper, 10 pt, conference]{ieeeconf}

\IEEEoverridecommandlockouts                              % This command is only needed if 
\usepackage{amsmath} % assumes amsmath package installed
\usepackage{amssymb}  % assumes amsmath package installed
\usepackage{pifont}%
\usepackage{flexisym}

\let\labelindent\relax
\usepackage{enumitem}

\usepackage[noadjust]{cite}
\setlist[itemize]{left=0pt,labelindent=0pt,labelsep=0.45em,itemsep=0pt,topsep=2pt}

\usepackage[table,dvipsnames]{xcolor}

\usepackage{graphicx}
\usepackage{diagbox}
\usepackage{booktabs}
\usepackage{multirow}
\usepackage{tabularx}
\usepackage{subcaption}
\usepackage{tikz}
\usepackage{cuted}

\usepackage[breaklinks,colorlinks]{hyperref}

\newcommand{\NA}[1]{$\spadesuit$\footnote{\color{red}Nikolay: #1}}

\title{\LARGE \bf
SplatSDF: Boosting SDF-NeRF via Architecture-Level Fusion with Gaussian Splats
}

\author{Runfa Blark Li$^{1}$, Daniel George$^{1}$, Keito Suzuki$^{1}$, Bang Du$^{1}$, K. M. Brian Lee$^{1}$, \\ Nikolay Atanasov$^{1}$, Truong Nguyen$^{1}$% <-this % stops a space
\thanks{This work was supported by the Ministry of Trade, Industry and Energy (MOTIE), Korea, under the Strategic Technology Development Program, supervised by the Korea Institute for Advancement of Technology (KIAT) [Grant No. P0026052].
}% <-this % stops a space
\thanks{Dept. of Electrical and Computer Engineering, University of California, San Diego (UCSD), La Jolla, CA, USA.
{\tt\scriptsize \{runfa, d1george, k3suzuki, b7du, kmblee, natanasov, tqn001\}@ucsd.edu} 
}%
}

\begin{document}

% 1. Render the title and footnotes (keeps \thanks visible)
\maketitle

% 2. Create the full-width figure immediately after
\begin{strip}
    \vspace*{-2.5cm}
    \centering
    % Adjust height/width as needed
    \includegraphics[width=\textwidth]{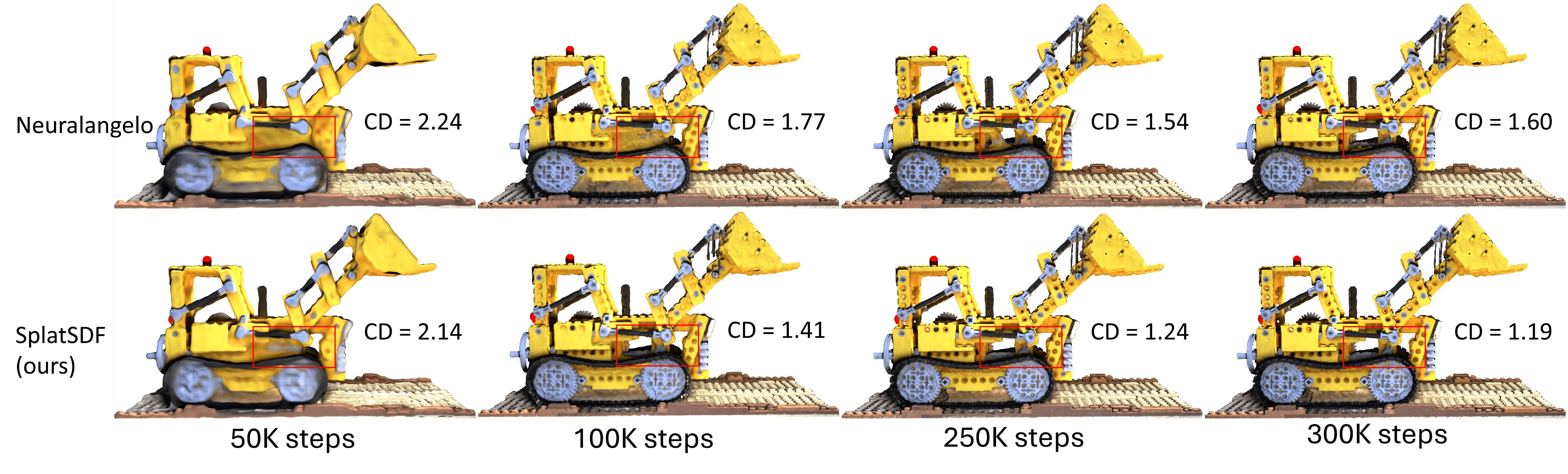}
    \captionof{figure}{SplatSDF accelerates SDF-NeRF training using a novel architecture that fuses 3D Gaussian Splat information. SplatSDF quickly caputres complex geometry, such as the holes in the red boxes, and converges $>3\times$ faster compared to the best baseline, Neuralangelo. SplatSDF achieves better chamfer distance (CD) at 100k epochs/3.97 hours than Neuralangelo at 300K epochs/15.15 hours. Visualization is available at: \url{https://blarklee.github.io/splatsdf/}.}
    \label{fig:first_fig}
\end{strip}

%%%%%%%%%%%%%%%%%%%%%%%%%%%%%%%%%%%%%%%%%%%%%%%%%%%%%%%%%%%%%%%%%%%%%%%%%%%%%%%%

\begin{abstract}
Signed distance-radiance field (SDF-NeRF) is a promising environment representation that offers both photorealistic rendering and geometric reasoning such as proximity queries for collision avoidance.
However, the slow training speed and convergence of SDF-NeRF hinder their use in practical robotic systems.
We propose SplatSDF, a novel SDF-NeRF architecture that accelerates convergence using 3D Gaussian splats (3DGS), which can be quickly pre-trained. 
% We explore the use of a pre-trained 3D Gaussian splat (3DGS) to accelerate the convergence of SDF-NeRF.
Unlike prior approaches that introduce a consistency loss between separate 3DGS and SDF-NeRF models, SplatSDF directly fuses 3DGS at an architectural level by consuming it as an input to SDF-NeRF during training.
This is achieved using a novel \emph{sparse 3DGS fusion} strategy that injects neural embeddings of 3DGS into SDF-NeRF around the object surface, while also permitting inference without 3DGS for minimal operation.
Experimental results show SplatSDF achieves 3$\times$ faster convergence to the same geometric accuracy than the best baseline, and outperforms state-of-the-art SDF-NeRF methods in terms of chamfer distance and peak signal to noise ratio, unlike consistency loss-based approaches that in fact provide limited gains.
We also present computational techniques for accelerating gradient and Hessian steps by $3\times$.
We expect these improvements will contribute to deploying SDF-NeRF on practical systems. 

% A signed distance function (SDF) is a useful representation for continuous-space geometry in many robotic operations, including geometric/photometric perception, and collision checking. Hence, reconstructing SDF from image observations accurately and efficiently is a fundamental problem. Recently, neural implicit SDF (SDF-NeRF) techniques, trained using volumetric rendering, have gained a lot of attention. Compared to earlier truncated SDF (TSDF) methods that fuse depth images on discrete voxels, SDF-NeRF enables continuous-space SDF reconstruction with better geometric and photometric accuracy. However, the accuracy and convergence speed of SDF reconstruction require further improvements. With the advent of 3D Gaussian Splatting (3DGS) as an explicit representation showing excellent rendering quality and speed, several works have focused on improving SDF-NeRF by introducing consistency losses on depth and surface normals between 3DGS and SDF-NeRF. However, loss-level connections alone lead to limited improvements. We propose a novel neural implicit SDF called ``SplatSDF'' to fuse 3DGS and SDF-NeRF at an architecture level with \emph{significant boosts to geometric and photometric accuracy and convergence speed}. Our SplatSDF relies on 3DGS as input \emph{only during training}, and keeps the same complexity and efficiency as the original SDF-NeRF during inference. Our method outperforms state-of-the-art SDF-NeRF models on geometric and photometric evaluation.
\end{abstract}  
\section{INTRODUCTION}

\label{sec:intro}
%\BL{Could start with a broader goal - e.g., achieving both photometric vs geometric accuracy.}
Signed distance fields (SDFs) are useful environment representation in robotics as they offer important geometric reasoning for environment reconstruction \cite{monoselfrecon}, and motion planning~\cite{bubble, shared-autonomy, MonoPLFlowNet, SM3D}
%capabilities such as proximity queries for collision avoidance in motion planning~\cite{bubble,curobo_report23}, information quantification in exploration~\cite{exploration} and surface normal queries for manipulation planning~\cite{manipulation}.
SDF-neural radiance field (SDF-NeRF) techniques further enable photorealistic rendering by ray-marching an opacity field derived from the SDF~\cite{volsdf,neus}.
We are interested in developing efficient training techniques that allow using SDF-NeRF as a unified representation for both geometric and photometric reasoning.

A crucial challenge to this end is computational efficiency. 
% \BL{A bit weak - Runfa, why do we not use SDF-NeRF as is?}
Because the photometric accuracy of SDF-NeRF derives from volumetric rendering with ray marching, many epochs are required to distinguish object surfaces from free space.
The ambiguity between object surface and free space can also lead to poor convergence with spurious `ghost' artifacts. 

We explore the use of 3D Gaussian splats (3DGS)~\cite{3DGS, waveletgs, dwtgs, dynagslam, gs_quantizer, coarse2fine, openhuman4d, openvoc} to accelerate the convergence of SDF-NeRFs.
3DGS can be trained expediently through rasterization, but falls short of providing proximity queries that are essential for robotics. 
% \BL{May have to reconcile Mac Schwager group's work.}

To combine the benefits of the two approaches, we propose \emph{SplatSDF}, a novel approach that fuses 3DGS into SDF-NeRF at training time for enhanced convergence.
Unlike concurrent work where 3DGS and SDF are separate models only linked through consistency losses~\cite{neusg,gsdf,3dgsr,pings,gaussianroom}, we fuse 3DGS into SDF-NeRF at an \emph{architectural level} by consuming 3DGS as input to SDF-NeRF during training to guide convergence. 
Meanwhile, 3DGS is no longer required at inference time, thus yielding a minimal representation that provides both geometric and photometric accuracies.
These benefits are realized through a novel \emph{sparse 3DGS fusion strategy} that injects an embedding of 3DGS into SDF-NeRF only at the object surface. 
% As a result, we propose \emph{SplatSDF}, a novel approach to incorporate 3DGS at an \emph{architectural level} of an SDF-NeRF mode. In our formulation, 3DGS is needed as an input at training time but not at inference time, the SDF-NeRF can be used independently. 

Our experimental results show that SplatSDF achieves 3$\times$ faster convergence for the same geometric accuracy, as shown in Figure~\ref{fig:first_fig}, and the converged results outperform state-of-the-art (SOTA) SDF-NeRF methods in terms of chamfer distance (CD) and peak signal-to-noise-ratio (PSNR). This is in contrast to other approaches that guide SDF using GS through consistency losses, which we found to actually achieve limited gains.
We also present computational methods that speed up gradient and Hessian computation by $3\times$.
Ablation studies demonstrate the effectiveness of using all 3DGS attributes, as opposed to treating 3DGS as a point cloud. 
In sum, our contributions are as follows:
\begin{itemize}
    \item We propose \emph{SplatSDF}, a novel SDF-NeRF architecture that uses GS to accelerate convergence.
    \item A novel sparse 3DGS fusion strategy effectively injects neural embedding of 3DGS into SDF-NeRF.
    \item Experimental results show that SplatSDF actually improves the training speed and photometric/geometric accuracy of SDF-NeRF unlike prior approaches to GS fusion. 
\end{itemize}

\section{RELATED WORK}
% \iffalse
% The essence of training Neural Implicit SDF is to store the scene's SDF values in a SDF-MLP which takes 3D coordinates as input and output real SDF values. Based on training, 
% \fi
% We categorize neural implicit SDF into rendering-based, pointcloud-based, and 3DGS-driven.

% \noindent
% \textbf{Rendering-based Neural Implicit SDF.}
% \noindent
% \textbf{Point Cloud-based Neural Implicit SDF.}
To train SDF representations, the prevailing approach is to supervise them geometrically using point clouds rather than images.
DeepSDF~\cite{deepsdf}, widely considered as the original work on neural SDF, is trained from pure point clouds. 
\iffalse Later works improve upon it mostly by building novel losses.\fi 
NeuralPull \cite{neurallpull} proposes a ``pulling-loss'' to pull randomly sampled query points to the nearest point from the point cloud along the direction of SDF gradient. 
iSDF \cite{isdf} scales geometric supervision to real-time mapping by keyframe selection.
\iffalse \cite{gridpull} improves upon \cite{neurallpull} by implementing the ``pulling-loss'' in grids sampled by a KD-tree around the point cloud to speed up search for the nearest point. \fi %\cite{smallstep} designs a pseudo-SDF ground truth from point cloud for loss at each small step of training to guide the convergence and avoid local minima. 

We take inspiration from these methods and use 3DGS as a geometric cue, which is more accurate than raw pointclouds. However, unlike the approaches, our end-result of SDF-NeRF also provides photometric accuracy.

% However, these methods highly depend on the quality of the point cloud since the SDF-MLP can converge to an inaccurate surface with an inaccurate or sparse point cloud. The absence of images also makes them impossible to render photometric details like SDF-NeRF.
% Meanwhile, our approach uses 3DGS to not only capture photometric details for SDF-NeRF, but also geometric cues as is done by point cloud-based approaches.

% Another important branch of work uses pure point clouds to train an SDF-MLP without images. 
\iffalse Accurate and dense point cloud shares sufficient prior knowledge of object surfaces and surface normals, which are the key ingredients to guide SDF-MLP converged.\fi 

To achieve photometric accuracy, volumetric rendering is required.
Whereas NeRF \cite{nerf, glossgau, dwtnerf} uses volumetric rendering to train MLPs that store the color and opacity of a scene, an SDF-NeRF replaces the opacity MLP with an MLP for the SDF of a scene \cite{neus}. For supervision with images, volumetric rendering is achieved by converting the SDF to opacity using a density function, such as the logistic distribution~\cite{neus} or the Laplace distribution~\cite{volsdf}.
% We build on these approaches to supervise a SDF-MLP using images, 
% In principle, any unimodal (i.e. bell-shaped) density distribution centered at 0 can be adopted as the SDF density function.
\iffalse In practice, the most widely used distributions are the logistic density distribution \cite{neus} and the Laplace distribution \cite{volsdf}. Improving high-frequency details is important for geometry, where \cite{hfneus} proposes to gradually increase the frequency of the positional encoding. Other works explore different encoders as alternatives to positional encoding for high-frequency details, where \cite{petneus} uses Tri-plane with SDF-MLP, \cite{co-nir} proposed a coordinate quantization strategy, and \cite{neuralangelo} uses hash-grid encoder from \cite{instantngp}. \fi 

3DGS is a more recent photometric representation than NeRF that replaces volumetric rendering with rasterization for faster training.
Therefore, a possible alternative direction to ours is to improve the geometric capabilities of 3DGS.
In this regard, improvements of 3DGS, such as SUGAR~\cite{sugar}, Gaussian surfels~\cite{gaussiansurfels}, GOF~\cite{gaussianopaictyfield} and 2DGS~\cite{2DGS} introduce consistency losses that improve geometric accuracy in terms of depth rendering.
However, all of these methods require intermediate depth rendering for proximity queries or surface extraction, using methods such as Poisson surface reconstruction~\cite{sugar,gaussiansurfels}, truncated SDF fusion on a voxel grid~\cite{2DGS}, or marching tetrahedra~\cite{gaussianopaictyfield}.
We focus on training an SDF-NeRF, which directly allows continuous proximity queries that are useful for robotics without rendering depth. 

Some concurrent work also aims to leverage 3DGS for faster training of SDF-NeRF. These methods introduce a consistency loss between separate 3DGS and SDF-NeRF models, such as surface normal consistency in NeusG~\cite{neusg} and 3DGSR~\cite{3dgsr}, depth consistency in GSDF~\cite{gsdf}. GaussianRoom~\cite{gaussianroom} adds edge detection to surface normal consistency losses for indoor scenes.
In contrast to these methods, the proposed SplatSDF first pre-trains a 3DGS, and directly uses the pre-trained 3DGS as an input to SDF-NeRF at training time. 
This has two notable merits: we can exploit the fact that 3DGS can be pre-trained much quicker than SDF-NeRF, and the resulting performance significantly improves upon stand-alone SDF-NeRF, which, as our experiments show, is not the case with the consistency loss-based approaches.

\begin{figure*}[!t]
  \centering
  \includegraphics[width=\textwidth]{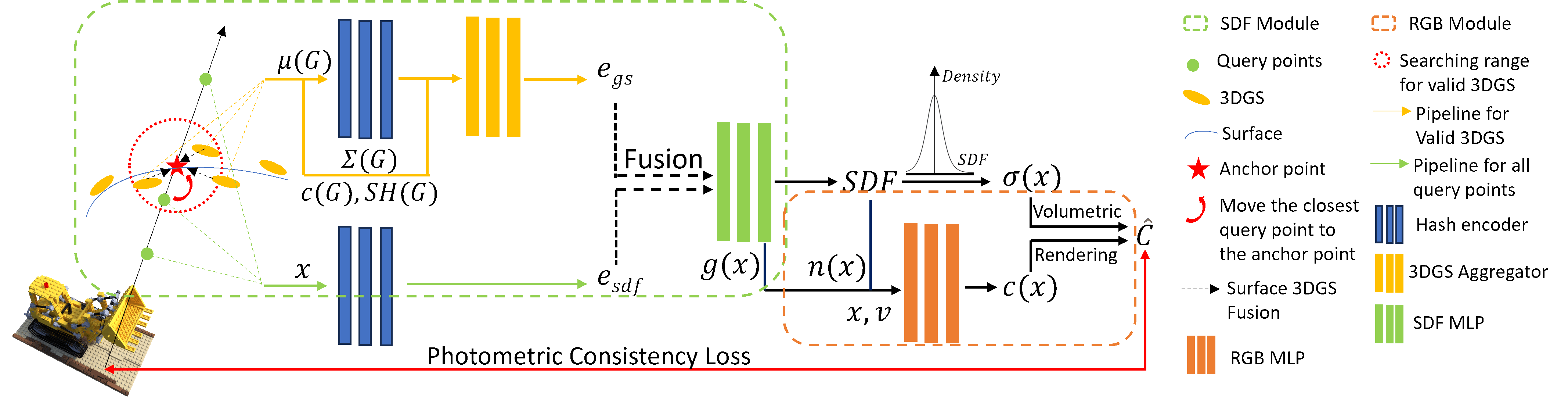}
  \caption{\textbf{Overview}. SplatSDF takes posed RGB images and 3DGS to train an SDF-NeRF model. We use 3DGS-rendered depth to identify the anchor point and shift the closest query point to the anchor point. With a shared hash encoder, we construct query-point SDF embeddings $e_{sdf}$ and 3DGS embeddings $e_{gs}$. Our method applies a \emph{3DGS aggregator} to merge the 3DGS attributes: mean $\mu$, covariance $\Sigma$, color $c$, and spherical harmonics $SH$. A novel \emph{surface 3DGS fusion} strategy merges $e_{gs}$ and $e_{sdf}$ only around anchor points to regress the SDF value. The SDF value is converted to per-point density $\sigma(x)$. We take the geometric features $g(x)$, the surface normal from SDF $n(x)$, the query-point coordinates $x$, and the viewing angle $v$ to estimate per-point color $c(x)$ and obtain the per-pixel color $\hat{C}$ by volumetric rendering, supervised with input images. Our contribution is the design of the \emph{3DGS aggregator} and the \emph{surface 3DGS fusion}.}
  \label{fig:overview}

\end{figure*}
%\NA{Make the use of subsections and bolded paragraphs consistent. For this section, I recommend having Problem Statement and SplatSDF Overview as regular subsections rather than bolded paragraphs.}

% \noindent
\section{PROBLEM STATEMENT} 
We assume that we are given RGB images with known camera poses and a sufficiently trained 3DGS \cite{3DGS} model $\mathcal{G}$ of the scene.
%, along with RGB images with known camera poses. 
Each Gaussian $G \in \mathcal{G}$ is parameterized by a mean $\mu(G)$, a covariance $\Sigma(G)$, opacity $\alpha(G)$, and spherical harmonics $SH(G)$.
The RGB images may be given or rendered from the 3DGS model $\mathcal{G}$.

Our objective is to recover an SDF-NeRF, comprising an SDF $f_{\mathcal{S}} : \mathbb{R}^{3} \to \mathbb{R}$ of the occupied space $\mathcal{S}$ in the scene, and a radiance field $\mathbf{c}(\mathbf{x}) : \mathbb{R}^{3} \to \mathbb{R}^{3}$ The SDF is
%\iffalse\BL{Here I am overloading notation slightly. Can we treat the scene as a set?} \RL{I checked it with GPT it says it's quite common treating the scene as a set}\fi, 
defined as $f_{\mathcal{S}}(\mathbf{x}) = \pm \inf_{\mathbf{y} \in \partial \mathcal{S}} \|\mathbf{x}-\mathbf{y}\|$ with positive sign when $\mathbf{x} \notin \mathcal{S}$ and negative sign otherwise. This allows geometric reconstruction of the surface of $\mathcal{S}$ as the zero-level set of its SDF $\partial \mathcal{S} = \left\{ \mathbf{x} \in \mathbb{R}^3 \mid f_{\mathcal{S}}(\mathbf{x}) = 0 \right\}$. Our approach uses the 3DGS $\mathcal{G}$ during training but yields the SDF $f_{\mathcal{S}}$ as an MLP that can be queried independently of $\mathcal{G}$ at inference time.

% Our goal is to train the function $f_{\mathcal{S}}(\mathbf{x})$ as a SDF-MLP. 
%\NA{We should also describe the input data and any supervision signal we assume available.}

% BL: Input vs output of the algorithm

% \subsection{SplatSDF}
\paragraph*{\textbf{Overview of approach}}
Figure \ref{fig:overview}
%\NA{In the caption we claim that both the 3DGS aggregator and the 3DGS fusion parts are our contributions but the description says very little about the 3DGS aggregator. Add a bit more about why merging the 3DGS attributes is important or novel.} 
shows an overview of the proposed SplatSDF.
%\NA{Perhaps, we should not use emph for these to not confuse the reader about our main contributions: 3DGS aggregator and 3DGS fusion.}. 
Our contribution focuses on the SDF, whereas the  radiance field is designed similarly to NeRF \cite{nerf}. 

A 3DGS aggregator~(Sec.~\ref{sec:aggregator}) constructs a 3DGS embedding $e_{gs}$, 
which is used in conjunction with an SDF embedding $e_{sdf}$ during training. Then, the per-point SDF is predicted as:
\begin{equation}
    f_{\mathcal{S}}(\mathbf{x}) %= F_{sdf}(x) 
    = f_{sdf}(Fuse(e_{sdf}(\mathbf{x}), e_{gs}(\mathbf{x}, \mathcal{G}))),
\label{sdfmlp_train}
\end{equation}
where $f_{sdf}$ is an MLP, and $Fuse$ is the sparse 3DGS fusion module (Sec.~\ref{sec:fusion}).

At inference time, the 3DGS embedding $e_{gs}$ is optional, and the SDF model can be queried as:
\begin{equation}
    f_{\mathcal{S}}(\mathbf{x}) %= F_{sdf}(x) 
    = f_{sdf}(e_{sdf}(\mathbf{x})).
\end{equation}
% where $Fuse$\
%NA{We should not introduce and emphasize too many different terms because the readers will get confused about the main point. We can mention the name 3DGS aggregator and highlight this here.} 
% is a novel fusion method we detail in Sec.~\ref{sec:fusion}, and $f_{sdf}$ is an MLP. 
%\NA{You can introduce the two subsections where you introduce the terms \emph{3DGS aggregator} and \emph{3DGS fusion}. Btw, we need a good introduction of the two components. Currently, the \emph{3DGS fusion} is introduced but not named correctly, while the \emph{3DGS aggregator} is not introduced.}. 
%
In what follows, we concretely define our main contributions in the sparse 3DGS fusion and the 3DGS aggregator modules.
The sparse 3DGS fusion module (Sec.~\ref{sec:fusion}) fuses the SDF embedding $e_{sdf}(\mathbf{x})$ and the 3DGS embedding $e_{gs}(\mathbf{x})$ near the geometry's surface.
The 3DGS aggregator (Sec.~\ref{sec:aggregator}) constructs per-Gaussian embeddings $e_{g}(G)$ that are combined into a 3DGS embedding $e_{gs}$ by the sparse weighted 3DGS fusion module.

% present our main contributions that instantiate the model defined in~\eqref{sdfmlp_train}, which are a) the 3DGS aggregator and b) the sparse weighted 3DGS fusion modules. 

% The sparse weighted 
% The per-Gaussian embeddings $e_{g}(G)$ are combined into a 3DGS embedding $e_{gs}(\mathbf{x}, \mathcal{G})$ through a w

% In what follows, we discuss our main contributions that define $Fuse$ in \eqref{sdfmlp_train}: the 3DGS aggregator and the 3DGS fusion methods. 

% In what follows, we discuss our main contributions about \emph{Fuse} in \eqref{sdfmlp_train}: \emph{3DGS aggregator \& fusion} %\NA{Consistently use the same two names, i.e., we should have \emph{3DGS aggregator} and \emph{3DGS fusion}. Do not take shortcuts. This is the main contribution!}.
% The core of SplatSDF is the 3DGS Aggregator and Fusion, which we will discuss in more detail. 

\section{SplatSDF}

\subsection{3DGS aggregator}\label{sec:aggregator}
The 3DGS aggregator constructs per-Gaussian embeddings $e_{g}(G)$
%\NA{The notation for this and $e_{gs}$ is too similar and their distinction is not introduced until eq. (3). We should potentially use different notation and also discuss the relationship.} 
by combining the attributes of each Gaussian. 
The 3DGS attributes are important because they reflect shape beyond the center points, as illustrated in our results (Table \ref{fig:ablation}).
The 3DGS aggregator aligns the feature spaces to provide an explicit notion of local geometry beyond sparse center positions. 
To ensure consistency, we share the same hash-encoder $h$ \cite{instantngp} between the SDF embedding $e_{sdf}$ and the Gaussian embedding $e_{g}$, so that they are defined as:
\begin{equation*}
\begin{aligned}
    e_{sdf}(\mathbf{x}) &= h(\mathbf{x}), \\
    e_{g}(G) &= f_{agg}(h(\mu(G)), \Sigma(G), c(G), SH(G)) ,
\end{aligned}
\end{equation*}
where $f_{agg}$ is an MLP, and $\mu, \Sigma, c, SH$ are the center coordinate, covariance, color and spherical harmonics parameters of $G$. 
Because the 3DGS aggregator
%\NA{We should be consistent with the capitalization of aggregator -- either always upper or always lower.} 
shares the same hash encoder $h$ between $e_{sdf}$ and $e_{g}$, the two embeddings remain consistent, and hence the 3DGS accelerates training but is not required at inference time. 
% in order to guarantee the consistency between GS \& SDF embeddings. 
% Since the SDF embeddings are guided by GS embeddings during training and memorized by the SDF-MLP,
%\NA{What is an SDF-MLP? This is the first time we mention it. Overall, the presentation is too "expert"-level so that someone who is not well versed with all the GS and SDF lingo will have a hard time following. We should explain the steps more clearly so that even someone working on motion planning can understand the main idea.}
% it is unnecessary to recall the GS embeddings during inference. 
We do not aggregate the GS opacity $\alpha$ in $f_{agg}$ at this stage, because it plays an important role in the following fusion step.

% \NA{Should this be $e_{gs}(G)$?}

\subsection{Sparse 3DGS fusion}\label{sec:fusion}
The sparse 3DGS fusion stage constructs 3DGS embedding $e_{gs}(\mathbf{x}, \mathcal{G})$ from the per-Gaussian embeddings $e_{g}(G)$, and fuses them with the SDF embeddings $e_{sdf}$. 
% to query-point Gaussian embeddings $e_{gs}(\mathbf{x}, \mathcal{G})$ %\NA{What does among themselves mean? Remove this, i.e., ``The 3DGS fusion stage combines the Gaussian embeddings $e_{gs}(G)$ and the query-point SDF embedding $e_{sdf}$ to implement the fusion function $Fuse(e_{sdf}(\mathbf{x}), e_{gs}(\mathbf{x}, \mathcal{G}))$ in \eqref{sdfmlp_train}.} \RL{Professor, I think Brian has made the correct annotations here, it should be $e_{g}(G)$, because $e_{g}(G)$ is the embedding of per-GS. To make it clear we decided to move the 3DGS aggregator section first, so that it's not confused where e_{gs} comes from, and I rewrote the sentence} 
% for boosting the query-point SDF embeddings $e_{sdf}$. 
This implements the fusion function $Fuse(e_{sdf}(\mathbf{x}), e_{gs}(\mathbf{x}, \mathcal{G}))$ in \eqref{sdfmlp_train}. 
% The section first discusses the Dense 3DGS Fusion strategy and its flaws and then explains our novel Surface 3DGS Fusion.

%\NA{Can we use paragraph* to have slightly better spacing? We can still have the title bold.}
\paragraph*{\textbf{Weighted Fusion of per-Gaussian Embeddings}} Inspired by the alpha blending in 3DGS~\cite{3DGS}, we propose a novel weighted blending strategy 
%\NA{Too many things are emphasized. It is starting to defeat the purpose of emphasis.} 
to fuse the per-Gaussian embeddings $e_{g}(G)$ into a GS embedding $e_{gs}(\mathbf{x}, \mathcal{G})$ at each query point $\mathbf{x}$, as:
\begin{equation}
    e_{gs}(\mathbf{x}, \mathcal{G}) = \frac{1}{K} \sum_{G \in \text{KNN}(\mathbf{x}, \mathcal{G})} \! e_{g}(G) w(\mathbf{x}, G) \alpha(G).
    % &w(\mathbf{x}, G) = \exp\left( -\frac{1}{2} (\mathbf{x} - \mu(G))^T\Sigma(G)^{-1}(\mathbf{x} - \mu(G)) \right)
\label{eq:fusion}
\end{equation}
%
% $w(\mathbf{x}, G) = \exp\left( -\frac{1}{2} (\mathbf{x} - \mu(G))^T\Sigma(G)^{-1}(\mathbf{x} - \mu(G)) \right)$
Here, $\text{KNN}(\mathbf{x}, \mathcal{G})$ denotes the $K$ GS with means nearest to the query point $\mathbf{x}$, and $\alpha(G)$ is the opacity of Gaussian $G$. The weight $w(\mathbf{x}, G)= \exp\left( -\frac{1}{2} (\mathbf{x} - \mu(G))^T\Sigma(G)^{-1}(\mathbf{x} - \mu(G)) \right)$ and opacity $\alpha$ balance the contribution of neighbor GS embeddings to the query-point GS embedding. Division by the number of selected Gaussians $K$ normalizes the resulting embedding. 

% While 3DGS uses a 2D Gaussian weight function for rendering based on the 2D distance between the GS and the center pixel, we design a 3D Gaussian weight function for the GS fusion based on the K-Nearest GS to the center 3D query point they belong to. The weight of the kth GS contributing to the center query point xij is defined as:

The weighted blending strategy is an extension of the 2D blending strategy in 3DGS~\cite{3DGS} to 3D in the following sense. 
While 3DGS uses a projected 2D Gaussian weight function for blending colors in the image plane, our approach uses the 3D Gaussian weight function to blend the embeddings in the 3D space directly.
Furthermore, the KNN algorithm can be seen as the 3D equivalent to frustum culling in the 2D projection step of 3DGS rasterization. Moreover, the original 3DGS uses 2D elliptical weighted average (EWA) volumetric splatting. Including opacity for 3D EWA is our novel design, which allows varying the contribution of each Gaussian based on its opacity in the scene. This is particularly useful for handling overlapping or less reliable Gaussians, as well as regions with varying densities. 
% In place of frustrum culling, we select the K-nearest Gaussians, since there is no frustrum in the 3D case. 
% This fusion strategy can be seen as the 3D analogue of the 2D image rendering in 3DGS~\cite{3DGS} in the following sense. 
% The weighted fusion is analogous to the alpha blending, except no ordering is required in the 3D case. 

In our implementation, we accelerated the KNN algorithm by hashing the GS $\mathcal{G}$ and the set of query points $\mathbf{X} = \{\mathbf{x} \}$ into $L^3$ voxels\iffalse \NA{We should use a different variable name. $\mathcal{S}$ was used to denote the scene occupancy.} \RL{Changed to $L$ now}\fi, inspired by PointNeRF~\cite{pointnerf}. Doing so decreases the computational complexity from $O( |\mathbf{X}| |\mathcal{G}|)$ in the naive case to $O(\frac{|\mathbf{X}||\mathcal{G}|}{L^3})$, where $|\mathbf{X}|$ and $|\mathcal{G}|$ are the number of query points and the number of Gaussians. Since we only consider Gaussians belonging to the same voxel as the query point, we are implicitly imposing a radius constraint in addition to the K-nearest requirement. %Further details are provided in the Supplementary Material. 

\paragraph*{\textbf{Surface 3DGS Fusion}} To fuse the GS embedding $e_{gs}(\mathbf{x}, \mathcal{G})$ with the SDF embedding $e_{sdf}(\mathbf{x})$, one possibility is to train an MLP that concatenates both GS and SDF embeddings and regresses another embedding of the same dimension. However, we present a fusion approach that is simpler but significantly more effective and efficient.

Our approach begins from the observation that all query points $\mathbf{x}$ lie along a ray $\mathbf{r}$ during training. We first compute an anchor point
%\NA{Reserve emphasis for the most important things that we really want the readers to remember.} 
$\mathbf{x}_{\mathbf{r}}$, which we define as the first intersection between the ray and the surface. This can be computed using the depth value rendered from the GS $\mathcal{G}$. Then, for each ray, we replace the closest query point to the anchor point $\mathbf{x}_{\mathbf{r}}$, with the anchor point $\mathbf{x}_{\mathbf{r}}$ itself, and take the GS embedding $e_{gs}(\mathbf{x}_{\mathbf{r}}, \mathcal{G})$ as the output. For all other points, we use the SDF embedding $e_{sdf}(\mathbf{x})$. Thus, we merely \emph{`replace'} the SDF embedding with the GS embedding at the anchor point, rather than \emph{`concatenating'} embeddings at all query points.

This fusion strategy is effective because the GS embedding is used only near the surface. This avoids spurious Gaussian blobs that are found further from the surface, which is a common problem in 3DGS. Moreover, the computational complexity is dramatically reduced compared to a dense fusion along the entire ray since we only compute the GS embedding at one anchor point per ray. 

\begin{figure}[t]
    \centering
    \includegraphics[width=\linewidth]{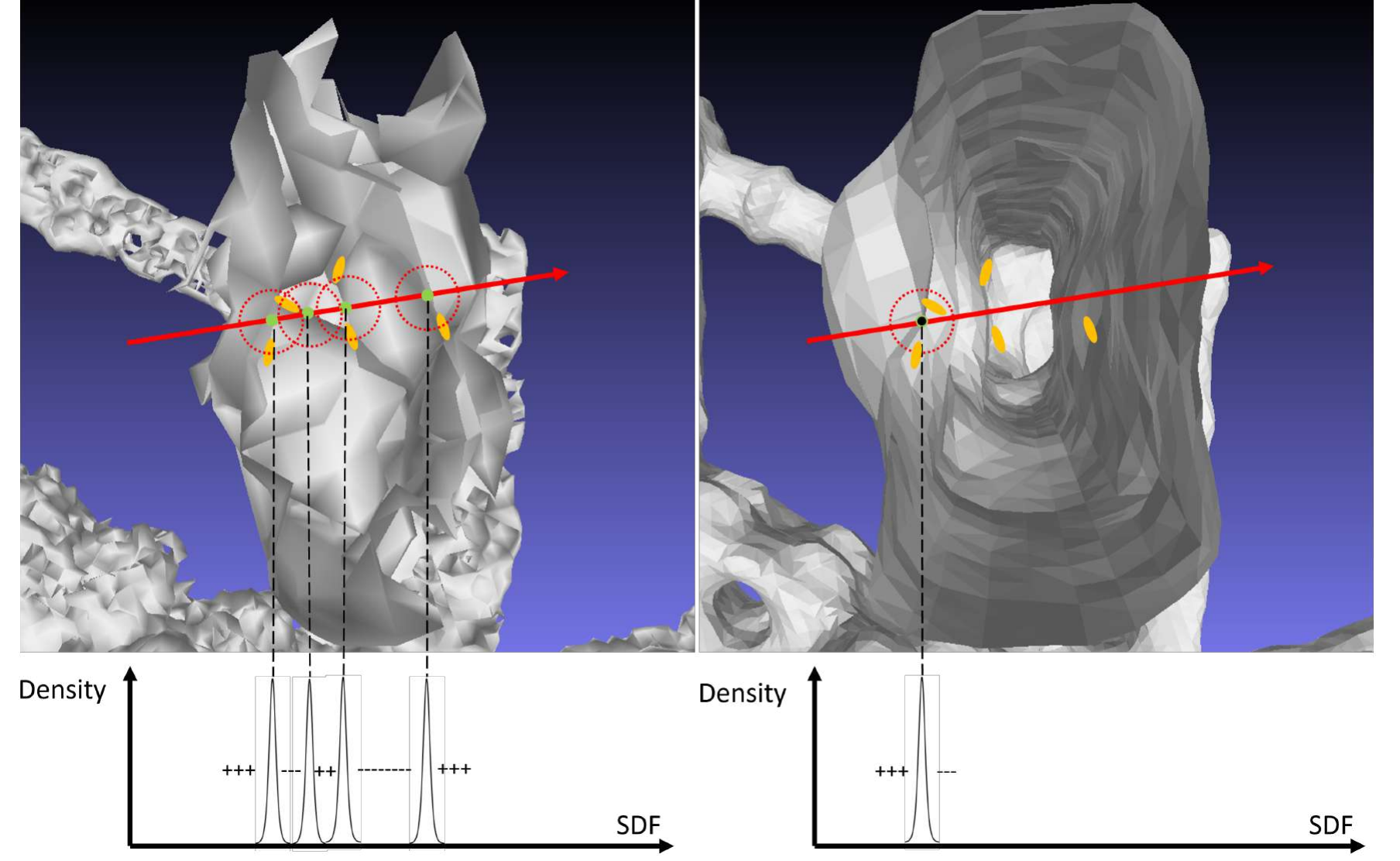}
    \caption{\textbf{Dense 3D Fusion vs Surface 3D Fusion}. Left: Dense 3DGS fusion on all valid query points (green points). Right: Surface 3DGS fusion only on the anchor point (black point). Fusing query points inside of surfaces using spurious GS (orange ellipsoids) far from the true surface leads to bumpy surface artifacts.}
    \label{fig:dense_vs_surface}
\end{figure}

This simple replacement with GS embeddings near the surface at the anchor points leads to notable improvements in convergence speed and accuracy over SDF-NeRF without GS.
% While densely fusing all valid query points is feasible for texture synthesis \cite{pointnerf}, it may not have the same effectiveness for geometry reconstruction. 
Figure~\ref{fig:dense_vs_surface} shows a qualitative comparison of the proposed strategy against a `dense' fusion via concatenation and MLP regression over both GS and SDF embeddings. 
The first row of Figure \ref{fig:dense_vs_surface} shows that the dense fusion approach (left) leads to bumpy surface artifacts. Further analysis shows that the zero-crossings of the SDF (i.e. peaks of density) align with spurious Gaussians, as shown in the second row.
This indicates that spurious Gaussians lead to errors in SDF.
% According to the density function in the second row, the zero-crossings of the SDF indicate where opacity $\sigma(\mathbf{x})$ is fused along the ray, 
%\iffalse \NA{How is this computed? What is the relationship with $\alpha$?} \RL{Added a few sentences here to help explain (Our experiment shows that flipping signs ... which proves that erroneous GS were fused along the ray). Please note that $\alpha$ is the GS opacity while $\sigma$ here is the query point opacity (SDF-NeRF per-point opacity)}\fi
% which proves that erroneous GS were fused to cause wrong SDF. 
Thus, using the GS embedding at query points far from the surface incorporates spurious Gaussians to the embedding, contributing erroneous density.
Although spurious Gaussians do not cause significant errors when projected to 2D images, they can cause errors in our SDF model as we fuse directly in 3D. Our sparse 3DGS fusion strategy mitigates this issue.

\subsection{Training}

Our SplatSDF model is trained for photometric accuracy by supervising volumetrically rendered images against the target images. 
To this end, we construct a radiance field that estimates the per query-point radiance $\mathbf{c}(\mathbf{x})$ using an RGB MLP $f_{color}$, whose inputs are the query position $\mathbf{x}$, viewing direction $\mathbf{v}$, geometric features $g(\mathbf{x})$, and the normals $\nabla_{\mathbf{x}} f_{\mathcal{S}}(\mathbf{x})$ (i.e., the SDF gradient): %\NA{I suggest removing "$\nabla_{\text{SDF}}(x)$" below, i.e., $n(x) = \nabla_{\text{SDF}}(x) = \frac{\partial sdf(x)}{\partial x}$. Also, it may be better to give a simpler name than $sdf(x)$ or at least do $\operatorname{sdf}(x)$.}
% For the ``Color module'', to regress per query-point radiance $c(x)$, a RGB-MLP $f_{color}$ is followed to take as input the query-points coordinates $x$, viewing direction $v$, query-points geometric features $g(x)$ at one layer before SDF-MLP regressing to SDF, and query-points normals $n(x)$ obtained from the gradients of query-points SDF:
\begin{equation}
\mathbf{c}(\mathbf{x}) = f_{color}(\mathbf{x},\mathbf{v},g(\mathbf{x}), \nabla_{\mathbf{x}} f_{\mathcal{S}}(x)).
\end{equation}
For simplicity, we do not explicitly input the GS $\mathcal{G}$ into the RGB MLP $f_{color}$ because the geometric features $g(\mathbf{x})$ and the normals $\nabla_{\mathbf{x}} f_{\mathcal{S}}(\mathbf{x})$ are already derived from the GS $\mathcal{G}$. 

For volumetric rendering, we convert the per-point SDF $f_{\mathcal{S}}(\mathbf{x})$ to per-point opacity $\sigma(\mathbf{x})$ using the logistic distribution $\phi_s(d) = se^{-sd}/(1+e^{-sd})^2$, where $s$ is the inverse of standard deviation, and $d$ is the SDF value at the query point $x$. The parameter $s$ is learnable, and is expected to approach zero as the SDF-MLP $f_{sdf}$ converges. The opacity is used in conjunction with the color module to volumetrically render images for supervision:
%\NA{There seem to be 1 too many parentheses below.}
%\iffalse\NA{How is $\sigma(\mathbf{x})$ related to $\phi_s(d)$?} \RL{Added the sentence "and $d$ is the SDF value at the query point $x$"}\fi 
%
\begin{equation}
\hat{\mathbf{C}}(\mathbf{r}) = \sum_{\mathbf{x}_{i} \in \mathbf{r}} T_i \left[1 - \exp\left(-\sigma\left(\mathbf{x}_{i}\right) \delta_i\right) \right] \mathbf{c}(\mathbf{x}_{i})
\label{volmetric_rendering},
\end{equation}
where $\delta_i = \|\mathbf{x}_{i+1} - \mathbf{x}_i\|_2$
%\NA{Write norm as $\|\cdot\|_2$.} 
is the distance between adjacent query points along the ray $\mathbf{r}$, and $T_i = \exp\left( -\sum_{j=1}^{i-1} \sigma_j \delta_j \right)$ denotes the accumulated transmittance.

%The latest GS-based surface reconstruction works \cite{sugar,gaussiansurfels,gaussianopaictyfield} and the concurrent GS-SDF works \cite{gsdf,3dgsr,gaussianroom} adopt new losses from geometric priors such as GS surface normals and depth. 

Our method is trained with the same losses as Neuralangelo \cite{neuralangelo}: L1 photometric loss, Eikonal loss, and curvature loss. To show the effectiveness of our architecture, we do not use any auxiliary losses with depth or normal priors.

\section{EXPERIMENTS}
\subsection{Datasets \& Implementation Details}
We use the DTU \cite{dtu} and NeRF Synthetic datasets \cite{nerf} for training and evaluation. We use 12 scenes from the DTU dataset which contains either 49 or 64 posed images per-scene obtained by a robot-held monocular RGB camera and the point cloud ground truth obtained from a structured-light scanner. For the NeRF Synthetic Dataset, we use 5 objects with 100 posed images for each scene. We use the ``Lego'' scene for ablation study.

We train the 3DGS \cite{3DGS} with the GS centers fixed at input pointcloud to retain geometric fidelity.
% \BL{Does this mean the centers/means of the GS do not change during training?}
As a reminder, we fix the well-trained GS and do not jointly optimize with SDF-NeRF, unlike concurrent work that jointly optimizes SDF and GS \cite{gsdf,3dgsr}.
Separate training of GS and SDF is beneficial because we can exploit the fast training speed of GS independently of SDF, while the end result of SDF-NeRF is not only more geometrically accurate (i.e. low CD) than GS, but also as photometrically accurate (i.e. high PSNR) as GS.
% Joint optimization has no merit because 1. SDF-NeRF covers all functionalities of GS: accurate photometry (i.e. high PSNR) and geometry (i.e. low Chamfer Distance (Chamfer Distance)); 2. GS can be trained quickly and accurately without SDF-NeRF. 
Similarly, PointNeRF \cite{pointnerf} uses accurate pointcloud to boost NeRF, but not NeRF to improve the pointcloud. 

For our 3DGS aggregator, we use a 3-layer MLP where the 2nd layer concatenates the upper-triangle of the GS covariance and the 3rd layer concatenates the GS color and SH. We do not use object segmentation masks as in some previous works, but instead randomly sample 512 pixels for each view, and sample 128 points per ray. We follow Neuralangelo \cite{neuralangelo} to sample foreground and background points separately in the two SDF-MLP. We use uniform sampling for the background with 32 points and a coarse-to-fine sampling \cite{nerf} for the foreground with 96 points. We force the anchor points to be on the foreground by only implementing the surface GS fusion on the foreground NeRF.

We use $CD$ in $mm$ as the geometric evaluation metric. As per previous work, we sample 3D grid coordinates to estimate the SDF and use Marching Cubes to get the surface mesh. We then sample points from the mesh and compare with ground truth points to compute the $CD$. We use PSNR as the photometric evaluation metric. For the NeRF Dataset, we train on the standard training split with 100 images per scene and test on the standard testing split with 200 images per scene. For the DTU dataset, since there are no standard train/test splits, we train/test on the same images per-scene.

\iffalse
In this work, we do not compare the RGB synthesis (in PSNR) for two reasons: 1. We already have well-trained 3DGS as input to SplatSDF, the RGB synthesis can be obtained with 3DGS. 2. There is no advantage of rendering RGB with SDF-NeRFs, since the RGB-MLP must be used with the SDF-MLP, which leads to redundant computation compared to 3DGS even standard NeRFs.
\fi

\iffalse
The proposed numerical results of our SplatSDF are obtained by sampling SDF at the resolution of 512\textsuperscript{3}, instead of 2048\textsuperscript{3} that is used in Neuralangelo which takes long time inference on a single scene. By using a low resolution, the comparison favors a bit on other methods but we still outperform them all.
 \fi

\subsection{Qualitative Results}

\begin{figure}[t]
    \centering
    \includegraphics[width=\columnwidth]{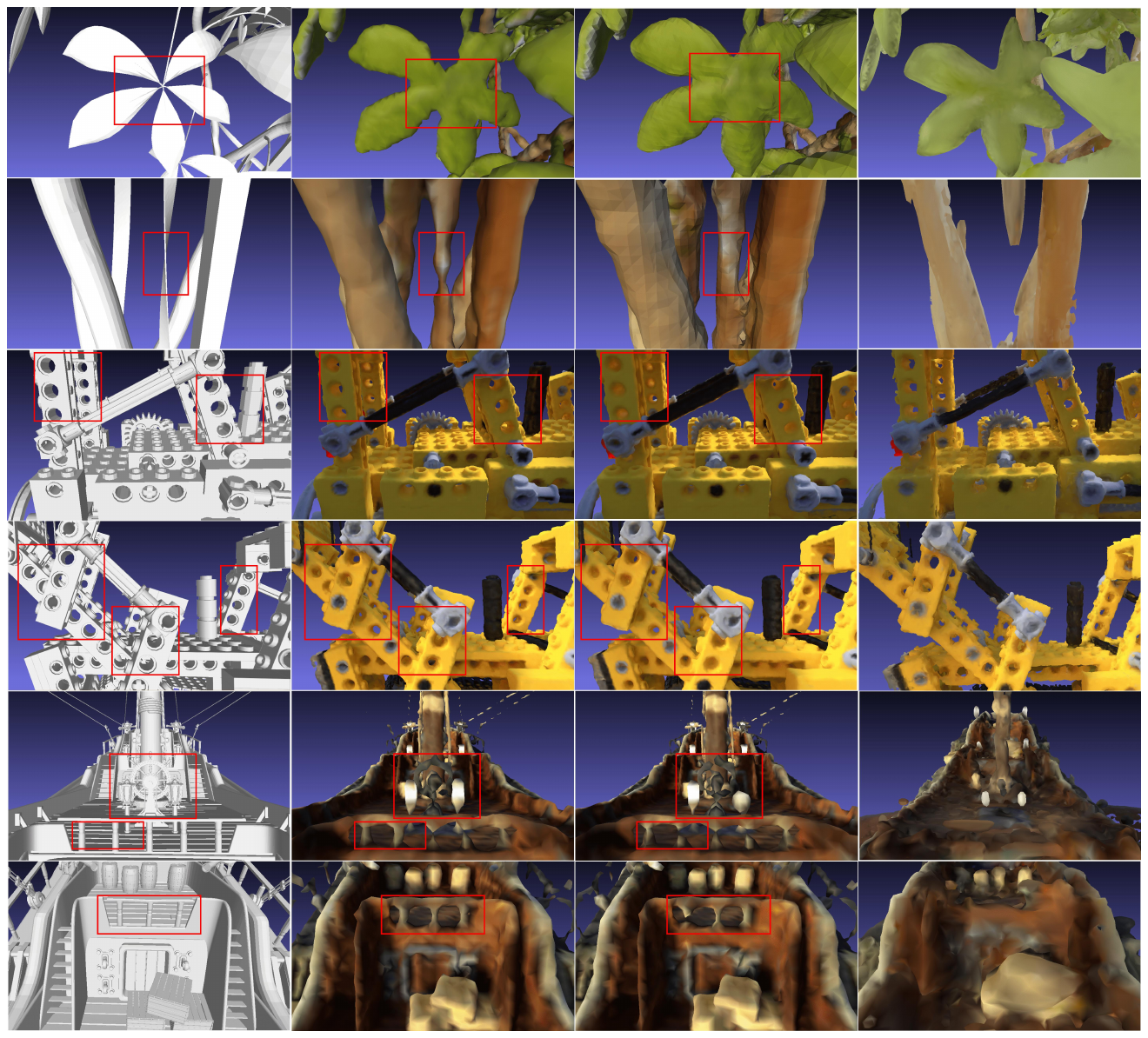}
    \caption{\textbf{Surface Mesh Comparison} on the NeRF Synthetic Dataset. Left to right: Ground Truth mesh (color is not available), SplatSDF, Neuralangelo, SuGAR. Row 1-2: Ficus. Row 3-4: Lego. Row 5-6: Ship. Zoom in to check details in red boxes. No red boxes for SuGAR since it is overall worse than SDF-NeRFs. We only show zoom-in details in this figure.}
    \label{fig:visual_comparison}
\end{figure}

\begin{figure}[t]
    \centering
    \includegraphics[width=\columnwidth]{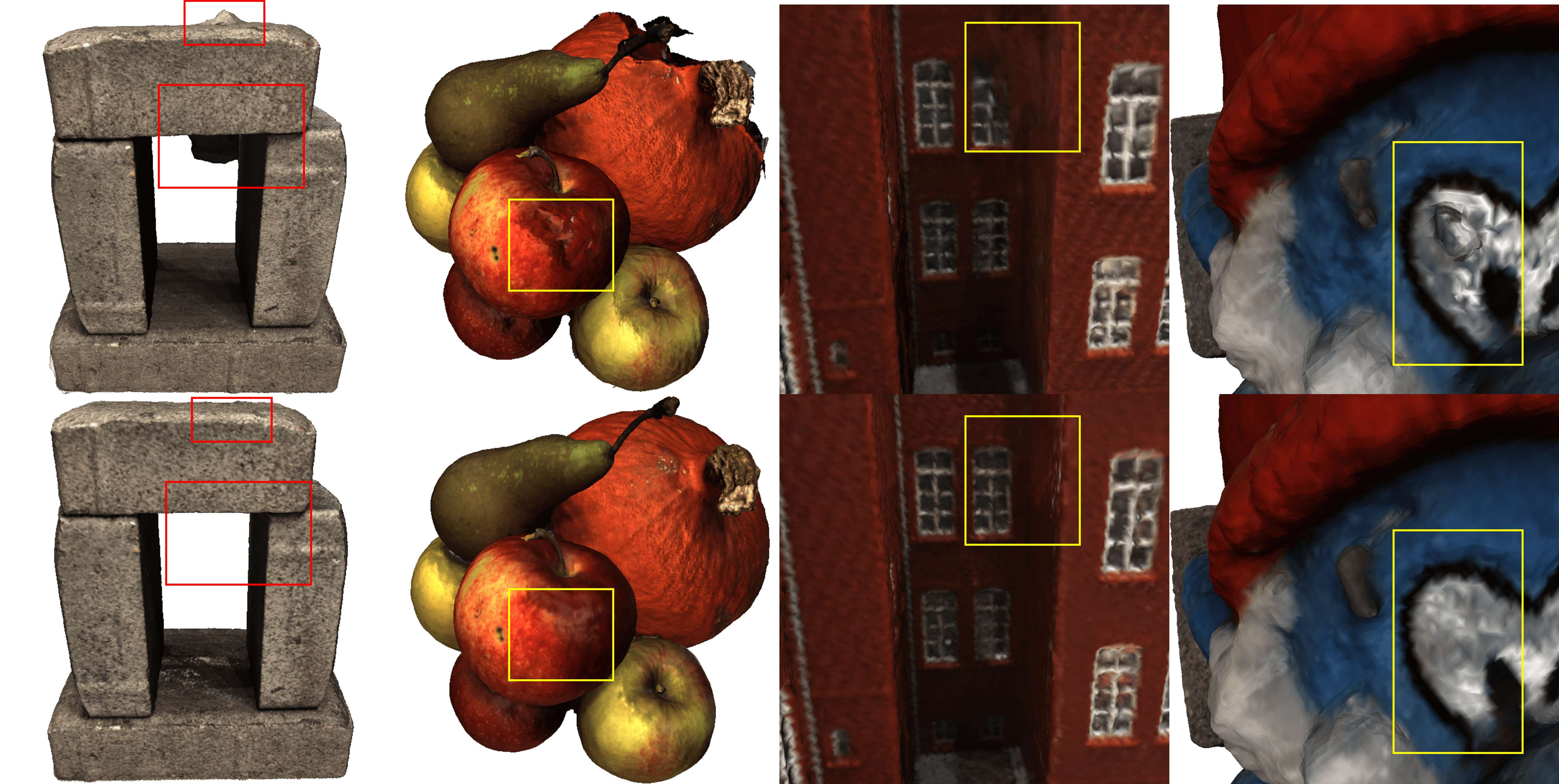}
    \caption{\textbf{Qualitative comparisons on the DTU Dataset.} Top row: Neuralangelo \cite{neuralangelo}. Bottom row: Our SplatSDF. We use the same models as Quantitative results from Table \ref{table:dtu_cd} to generate the visual results. Zoom in to check the details in the red or yellow bounding boxes.}
    \label{fig:dtu_vis}
\end{figure}

\begin{table*}[t]
\caption{\textbf{Quantitative Results on the DTU Dataset} (Chamfer Distance in mm $\downarrow$).\colorbox{yellow}{Yellow}is the best and \colorbox{pink}{pink}is the second best. Our SplatSDF achieves the best geometric accuracy. All results are from latest papers and we validate the results of the baseline Neuralangelo. See Supplementary for the source of each result.}\label{table:dtu_cd}
\resizebox{\linewidth}{!}{%
\begin{tabular}{cllllllllllllll}
& Scan ID          & 24   & 37   & 40   & 55   & 63   & 65   & 69   & 83   & 105  & 106  & 110  & 114  & Mean \\ \hline
\multirow{2}{*}{Traditional} & COLMAP \cite{colmap}          & 0.81 & 2.05 & 0.73 & 1.22 & 1.79 & 1.58 & 1.02 & 3.05 & 2.05 & 1.00 & 1.32 & 0.49 & 1.43 \\
                             & NeRF \cite{nerf}            & 1.90 & 1.60 & 1.85 & 0.58 & 2.28 & 1.27 & 1.47 & 1.67 & 1.07 & 0.88 & 2.53 & 1.06 & 1.51 \\ \hline
\multirow{16}{*}{\begin{tabular}[c]{@{}c@{}}SDF-NeRF\end{tabular}} & UNISURF \cite{unisurf}         & 1.32 & 1.36 & 1.72 & 0.44 & 1.35 & 0.79 & 0.80 & 1.49 & 0.89 & 0.59 & 1.47 & 0.46 & 1.06 \\
                             & MVSDF \cite{mvsdf}           & 0.83 & 1.76 & 0.88 & 0.44 & 1.11 & 0.90 & 0.75 & 1.26 & 1.35 & 0.87 & 0.84 & 0.34 & 0.94 \\
                             & VolSDF \cite{volsdf}          & 1.14 & 1.26 & 0.81 & 0.49 & 1.25 & 0.70 & 0.72 & 1.29 & 0.70 & 0.66 & 1.08 & 0.42 & 0.88 \\
                             & NeuS \cite{neus}            & 1.00 & 1.37 & 0.93 & 0.43 & 1.10 & 0.65 & 0.57 & 1.48 & 0.83 & 0.52 & 1.20 & 0.35 & 0.87 \\
                             & NeuS-12 \cite{neus}         & 0.93 & 1.07 & 0.81 & 0.38 & 1.02 & 0.60 & 0.58 & 1.43 & 0.78 & 0.57 & 1.16 & 0.35 & 0.81 \\
                             & HF-NeuS \cite{hfneus}         & 0.76 & 1.32 & 0.70 & 0.39 & 1.06 & 0.63 & 0.63 & \cellcolor{yellow}1.15 & 0.80 & 0.52 & 1.22 & 0.33 & 0.79 \\
                             & MonoSDF \cite{MonoSDF}         & 0.66 & 0.88 & 0.43 & 0.40 & 0.87 & 0.78 & 0.81 & 1.23 & \cellcolor{pink}0.66 & 0.66 & 0.96 & 0.41 & 0.73 \\
                             & RegSDF \cite{regsdf}          & 0.60 & 1.41 & 0.64 & 0.43 & 1.34 & 0.62 & 0.60 & 0.90 & 1.02 & 0.60 & \cellcolor{yellow}0.59 & \cellcolor{yellow}0.30 & 0.75 \\
                             & PET-NeuS \cite{petneus}        & 0.56 & 0.75 & 0.68 & 0.36 & 0.87 & 0.76 & 0.69 & 1.33 & \cellcolor{pink}0.66 & \cellcolor{pink}0.51 & 1.04 & 0.34 & 0.71 \\
                             & OAV \cite{oav}             & 1.92 & 2.35 & 1.96 & 1.11 & 1.83 & 2.01 & 1.30 & 1.53 & 1.50 & 0.71 & 1.56 & 0.83 & 1.55 \\
                             & TUVR \cite{tuvr}            & 0.83 & 1.06 & 0.57 & 0.40 & 1.00 & 0.62 & 0.62 & 1.41 & 0.94 & 0.57 & 1.07 & 0.35 & 0.79 \\
                             & DebSDF \cite{debsdf2024}          & 0.71 & 0.94 & 0.46 & 0.39 & 1.05 & 0.61 & 0.59 & 1.49 & 0.88 & 0.61 & 1.05 & 0.34 & 0.76 \\
                             & NeuralWarp \cite{neuralwarp}      & 0.49 & 0.71 & 0.38 & 0.38 & \cellcolor{yellow}0.79 & 0.81 & 0.82 & 1.20 & 0.68 & 0.66 & \cellcolor{pink}0.74 & 0.41 & 0.67 \\
                             & Geo-NeuS \cite{geoneus}        & 0.46 & 0.85 & 0.38 & 0.43 & 0.89 & \cellcolor{yellow}0.50 & \cellcolor{yellow}0.50 & 1.26 & \cellcolor{pink}0.66 & 0.52 & 0.82 & \cellcolor{pink}0.31 & 0.63 \\
                             & Neuralangelo \cite{neuralangelo}    & \cellcolor{pink}0.37 & \cellcolor{pink}0.72 & \cellcolor{pink}0.35 & \cellcolor{pink}0.35 & 0.87 & \cellcolor{pink}0.54 & \cellcolor{pink}0.53 & 1.29 & 0.73 & \cellcolor{yellow}0.47 & \cellcolor{pink}0.74 & 0.32 & \cellcolor{pink}0.61 \\
                             & SplatSDF (ours)            & \cellcolor{yellow}0.35 & \cellcolor{yellow}0.67 & \cellcolor{yellow}0.32 & \cellcolor{yellow}0.31 & \cellcolor{pink}0.84 & \cellcolor{pink}0.54 & 0.55 & \cellcolor{pink}1.16 & \cellcolor{yellow}0.65 & \cellcolor{yellow}0.47 & 0.75 & \cellcolor{pink}0.31 & \cellcolor{yellow}0.58 \\ \hline
\multirow{6}{*}{\begin{tabular}[c]{@{}c@{}}GS-based \\ Surface \\ Reconstruction\end{tabular}} & Scaffold-GS \cite{scaffoldgs}     & 7.23 & 6.23 & 6.48 & 7.44 & 8.17 & 4.27 & 5.78 & 5.45 & 6.36 & 5.05 & 5.95 & 6.32 & 6.23 \\
                             & 3DGS \cite{3DGS}            & 2.14 & 1.53 & 2.08 & 1.68 & 3.49 & 2.21 & 1.43 & 2.07 & 1.75 & 1.79 & 2.55 & 1.53 & 2.02 \\
                             & SuGAR \cite{sugar}           & 1.47 & 1.33 & 1.13 & 0.61 & 2.25 & 1.71 & 1.15 & 1.63 & 1.07 & 0.79 & 2.45 & 0.98 & 1.38 \\
                             & GaussianSurfels \cite{gaussiansurfels} & 0.66 & 0.93 & 0.54 & 0.41 & 1.06 & 1.14 & 0.85 & 1.29 & 0.79 & 0.82 & 1.58 & 0.45 & 0.88 \\
                             & 2DGS \cite{2DGS}            & 0.48 & 0.91 & 0.39 & 0.39 & 1.01 & 0.83 & 0.81 & 1.36 & 0.76 & 0.70 & 1.40 & 0.40 & 0.79 \\
                             & GausianField \cite{gaussianopaictyfield}    & 0.50 & 0.82 & 0.37 & 0.37 & 1.12 & 0.74 & 0.73 & 1.18 & 0.68 & 0.77 & 0.90 & 0.42 & 0.72 \\ \hline
\multirow{2}{*}{\begin{tabular}[c]{@{}c@{}}GS-guided\\ SDF-NeRF\end{tabular}} & 3DGSR \cite{3dgsr}           & 0.68 & 0.84 & 0.70 & 0.39 & 1.16 & 0.87 & 0.77 & 1.48 & 0.87 & 0.69 & 0.80 & 0.42 & 0.81 \\
                             & GSDF \cite{gsdf}            & 0.59 & 0.94 & 0.46 & 0.38 & 1.30 & 0.77 & 0.73 & 1.59 & 0.76 & 0.59 & 1.22 & 0.38 & 0.81
\end{tabular}}
\end{table*}

We present qualitative comparisons against Neuralangelo \cite{neuralangelo}, which we found to be the strongest baseline. 
Figure \ref{fig:first_fig} shows that our SplatSDF achieves faster and more accurate convergence. SplatSDF only takes 100K steps to get $CD=1.41$, indicating $>3$ times faster convergence compared to Neuralangelo which takes 300K steps to get $CD=1.60$. SplatSDF takes 3.97 hours for 100K steps, which already achieves better CD than Neuralangelo trained to 300K steps taking 15.15 hours. 
SplatSDF also achieves better accuracy at convergence, capturing difficult shapes and details. While SDF-NeRF's iso-surface is always initialized as a unit sphere and ``pulled'' inside to fit to the convex surface, it is common to see it under-fitting to concave surfaces with small/thin details. This is because using the visual momentum alone from previous methods tends to smooth and blur the iso-surface and is insufficient to ``pull'' it to complex shapes. However, our SplatSDF amends this under-fitting with the novel architecture-level GS fusion. An example is shown in the red box in Figure \ref{fig:first_fig} where our model quickly captures the holes, whereas previous methods do not. All results in Figure \ref{fig:visual_comparison} are trained to converge and the details in the red boxes validate the improvements, such as the thin leaves and stems, the small holes in the Lego, the helm, the white lamp and the rail in the ship. We also include a comparison with the SOTA 3DGS-based surface reconstruction work SuGAR \cite{sugar}. Although SuGAR achieves faster surface reconstruction, its quality is lower than SOTA SDF-NeRFs since they transform the shape of the Gaussian primitives to fit the surfaces causing noisy artifacts and missing details. Moreover, it cannot estimate the distance field for arbitrary 3D coordinates. Our superior results on DTU as shown in Figure \ref{fig:dtu_vis} further validates the performance on real dataset, we show the colored mesh, not the rendered RGB view, the defects of Neuralangelo directly shows the geometric error instead of the photometric one.
% Regarding the comparison to the fastest Neural SDF NeuS2 \cite{neus2}, we discuss in Sec. \ref{sec:speed}. 

\begin{figure}[t]
    \centering
    \includegraphics[width=\columnwidth]{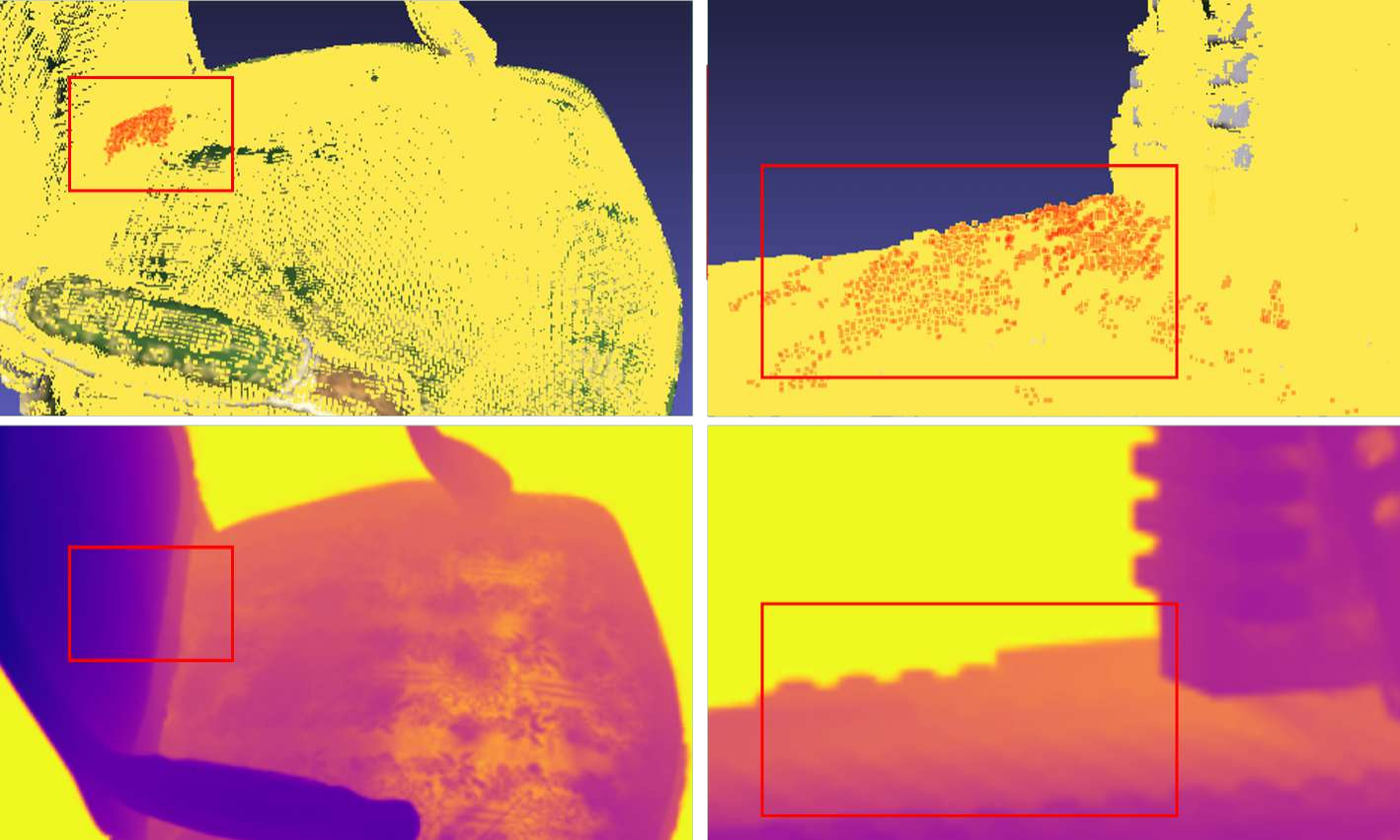}
    \caption{\textbf{Tolerance to erroneous 3DGS initialized from noisy point cloud}. First row: GS center in yellow overlap with the estimated surface mesh. Noisy GS centers is in red and in red boxes. Second row: No noise in the red box from rendered depth.}
    \label{fig:tolerance}
\end{figure}

\noindent
\textbf{Tolerance to noisy initialization} Figure \ref{fig:tolerance} shows that our SplatSDF can tolerate errors and noise in GS. We attribute this robustness to two reasons: 1. As shown in the second row of Figure \ref{fig:tolerance}, GS-rendered depth eliminates errors from erroneous GS centers because we use volumetric rendering rather than surface rendering and sstimate accurate anchor points to fuse correct surface 3DGS; 2. SDF-NeRF itself tends to under-fit to complex shapes by smoothing and blurring details, which alleviates the errors introduced by noisy 3DGS centers.

\subsection{Quantitative Results}
\label{sec:quantiative_results}

\begin{table}[t]
\caption{\textbf{Quantitative Results on NeRF Synthetic Dataset}. The top part shows the geometric accuracy in Chamfer Distance in mm $\downarrow$, and the bottom part shows the photometric accuracy in PSNR $\uparrow$. \colorbox{yellow}{Yellow} is the best and \colorbox{pink}{pink} is the second best. Our SplatSDF achieves the best overall accuracy.}\label{table:nerfsyn_cd}
\resizebox{\linewidth}{!}{%
\begin{tabular}{lcccccc}
                                                               & Chair & Ficus & Lego  & Mic  & Ship & Mean \\ \hline
VolSDF \cite{volsdf}                                                         & 1.18  & 3.01  & 2.26       & 1.13 & 6.42 & 2.80 \\
NeuS \cite{neus}                                                           & 3.99  & 0.94  & 2.56       & 1.00 & 5.38 & 2.77 \\
NeRO \cite{nero}                                                           & 1.27  & 1.22  & 1.90     & 0.87 & 4.95 & 3.72 \\
BakedSDF \cite{bakedsdf}                                                       & 1.83  & 10.9  & \cellcolor{pink}1.13      & 0.84 & 3.88 & 3.32 \\
NeRF2Mesh \cite{nerf2mesh}                                                      & 1.62  & \cellcolor{yellow}0.65  & 1.93       & 0.78 & 2.20 & 1.44 \\
\begin{tabular}[c]{@{}l@{}}RelightableG \cite{relightablegaussian}\end{tabular} & 3.65  & 1.26  & 1.63      & 1.76 & 3.35 & 2.33 \\
HF-NeuS \cite{hfneus}                                                   & \cellcolor{pink}0.69  & 1.12  & \cellcolor{yellow}0.94       & \cellcolor{yellow}0.72 & 2.18 & 1.13 \\
3DGSR \cite{3dgsr}                                                          & 1.01  & \cellcolor{pink}0.69  & 1.35       & 1.15 & 3.35 & 1.51 \\
Neuralangelo \cite{neuralangelo}                                                   & \cellcolor{yellow}0.56  & 1.12  & 1.60       & 0.78 & \cellcolor{pink}0.78 & \cellcolor{pink}0.97 \\
\begin{tabular}[c]{@{}l@{}}SplatSDF (ours)\end{tabular}      & 0.71  & 0.91  & 1.19      & \cellcolor{pink}0.75 & \cellcolor{yellow}0.72 & \cellcolor{yellow}0.86   \\ \hline
NeRF \cite{nerf}        & 33.00 & 30.15 & 32.54 & 32.91 & 28.34 & 31.39 \\
VolSDF \cite{volsdf}      & 25.91 & 24.41 & 26.99 & 29.46 & 25.65 & 26.48 \\
NeuS \cite{neus}        & 27.95 & 25.79 & 29.85 & 29.89 & 25.46 & 27.79 \\
HF-NeuS \cite{hfneus}     & 28.69 & 26.46 & 30.72 & 30.35 & 25.87 & 28.42 \\
PET-NeuS \cite{petneus}     & 29.57 & 27.39 & 32.40 & 33.08 & 26.83 & 29.85 \\
NeRO \cite{nero}        & 28.74 & 28.38 & 25.66 & 28.64 & 26.55 & 27.60 \\
BakedSDF \cite{bakedsdf}    & 31.65 & 26.33 & 32.69 & 31.52 & 27.55 & 29.95 \\
NeRF2Mesh \cite{nerf2mesh}   & 34.25 & 30.08 & 34.90 & 32.63 & 29.47 & 32.27 \\
Mip-NeRF \cite{mipnerf}   & \cellcolor{pink}35.14 & 33.29 & 35.70 & 36.51 & 30.41 & 34.21 \\
3DGS \cite{3DGS}        & \cellcolor{yellow}35.36 & 34.87 & \cellcolor{pink}35.78 & 35.36 & 30.80 & 34.43 \\
Ins-NGP \cite{instantngp}     & 35.00 & 33.51 & \cellcolor{yellow}36.39 & 36.22 & 31.10 & \cellcolor{pink}34.44 \\
Neuralangelo \cite{neuralangelo} & 34.72 & \cellcolor{pink}35.91 & 33.20 & \cellcolor{pink}36.79 & \cellcolor{pink}31.45 & 34.41 \\
SplatSDF (ours)     & 34.73 & \cellcolor{yellow}36.15 & 33.24 & \cellcolor{yellow}37.06 & \cellcolor{yellow}31.47 & \cellcolor{yellow}34.53
\end{tabular}}
\end{table}

 Table \ref{table:dtu_cd} shows a comparison on the DTU dataset over three categories. Our model outperforms the best baseline Neuralangelo \cite{neuralangelo} and achieves the lowest $CD$ over all previous works. We analyze two reasons why our method performs worse than the baseline on a few scenes. The first reason is that the ``anchor points'' estimated from 3DGS rendered depth are inaccurate in some areas and the 3DGS quality is insufficient. The second reason is that the DTU dataset contains erroneous point clouds in some regions. The 3DGS-based surface reconstruction methods generally train faster but their accuracy is lower than SOTA SDF-NeRF methods, as can be seen in Table \ref{table:dtu_cd} and Figure \ref{fig:visual_comparison}. We also show the results from two concurrent methods, GSDF \cite{gsdf} and 3DGSR \cite{3dgsr}, which use 3DGS to guide SDF-NeRF. Their results show that loss-level connections alone do not provide improvements over SDF-NeRF, where our architecture-level fusion method does. We conduct more comparisons on the NeRF Synthetic Dataset (Table \ref{table:nerfsyn_cd}), and the results show that our SplatSDF outperforms SOTA methods on both geometric and photometric accuracy, especially our baseline Neuralangelo~\cite{neuralangelo}. The results in Table \ref{table:nerfsyn_cd} further show that our model improves SDF-NeRF not only on geometric accuracy, but also on photometric accuracy. %Photometric evaluation (PSNR) on the DTU dataset is in Supplementary.

\subsection{Ablation Study}
\begin{figure}[t]
    \centering
    \includegraphics[width=\columnwidth]{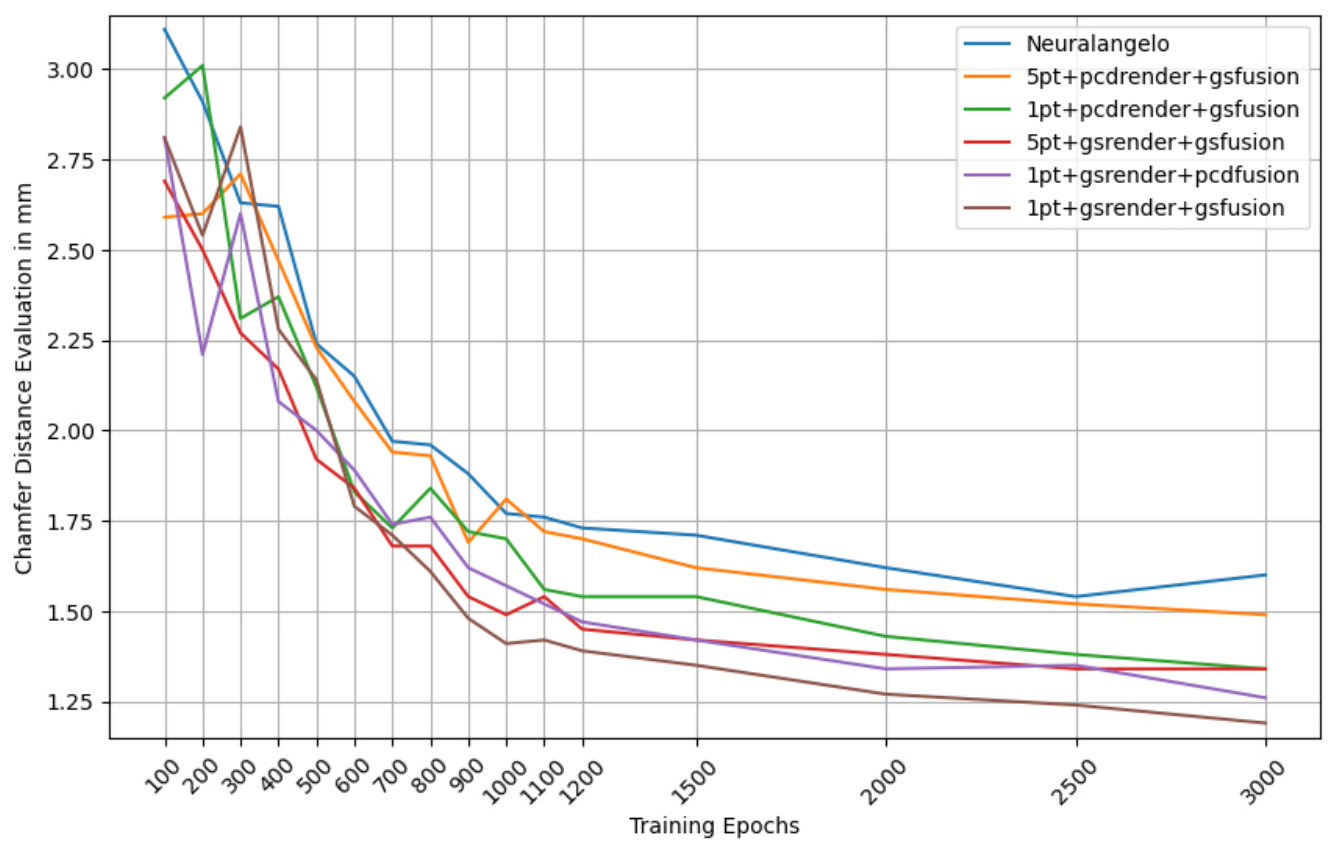}
    \caption{\textbf{Ablation study of geometric accuracy on ``Lego''}. All variations of our methods achieve faster convergence to better accuracy than the baseline ``Neuralangelo''.}
    \label{fig:ablation}
\end{figure}

We conduct ablation studies (Figure \ref{fig:ablation}) on the influence of three factors - sparse GS embedding fusion, the use of GS for depth images, and the use of GS over point clouds. All ablation variants outperform the strongest baseline, Neuralangelo, which shows the benefit of fusing pre-trained GS. 

Firstly, to examine the importance of Surface Fusion over Dense Fusion, we consider fusing five closest query points to the anchor point (5pt variants, shown in orange and red in Figure~\ref{fig:ablation}) as opposed to fusing only the anchor point (1pt variants, shown in green and brown in Figure~\ref{fig:ablation}).
% Compare all the “5pt” with its corresponding “1pt”, which are “orange VS green” and “red VS brown”. 
The results in Figure~\ref{fig:ablation} show that fusing only the anchor point as proposed outperforms fusing more query points near the anchor point, consistent with the observation in Figure \ref{fig:dense_vs_surface}. 

Secondly, we compare deriving anchor points based on GS-rendered depth (``gsrender'', shown in orange and green in Figure~\ref{fig:ablation}) and pointcloud-rendered depth (``pcdrender'', shown in red and brown in Figure~\ref{fig:ablation}).
% using GS rendered depth outperforms using point cloud rendered depth to find anchor points. Compare all the “pcdrender” VS “gsrender”, which are “orange VS red” and “green VS brown”. 
It can be seen from Figure~\ref{fig:ablation} that using GS-rendered depth outperforms the pointcloud-rendered depth.
This is because the depth point cloud from MVSNet is noisy, which leads to inaccurate surface anchor points. 
In contrast, the depth image from a well-trained GS accurately estimates the anchor points. 
This is consistent with the observation that SplatSDF can tolerate noisy point cloud in Figure \ref{fig:tolerance}. 

Lastly, we compare the proposed GS fusion (``gsfusion'', shown in purple in Figure~\ref{fig:ablation}) against point cloud fusion (``pcdfusion'', shown in brown in Figure~\ref{fig:ablation}).
The point-cloud fusion baseline is implemented by removing the GS covariance and SH features from the GS aggregator, and removing the GS covariance and opacity terms in the distance weight for 3DGS fusion.
The results in Figure~\ref{fig:ablation} shows that the proposed GS fusion approach outperforms pointcloud fusion, which illustrates the importance of using all GS features, including covariance, opacity and SH coefficients. 

\subsection{Training Acceleration}
\label{sec:speed}
%\NA{Can we call this section "Training Acceleration" instead of "Speedup"? The latter is rather informal.}

Fast training is important for robotics applications.
Our proposed SplatSDF reduces the steps for convergence. Here, we investigate how to accelerate individual steps. % in SplatSDF for the overall training time.

We found the main bottleneck to be the computation of first- and second-order derivatives (Gradient \& Hessian (GH), Table~\ref{tab:speed}). To accelerate the derivative computation, we first followed NeuS2~\cite{neus2} - the fastest implementation available. NeuS2 attributes the speedup to the use of a closed-form second-order approximation using TinyCUDANN (TCNN) for the backpropagation of SDF and RGB MLPs. We integrated TCNN from NeuS2 into our PyTorch for the SDF and RGB MLPs in both forward \& backward with NeuS2's purely CUDA C++. However, when testing over 1000 iterations on a 3090 Ti GPU (Table~\ref{tab:speed}), we found the speeds are nearly identical under the same architecture (19.95\,ms to 18.64\,ms), in contrary to the two orders of magnitude acceleration claimed in~\cite{neus2}. Moreover, using TCNN as MLP implementation also yields minor gains in forward steps (SDF: 3.19$\to$3.12 ms; RGB: 1.13$\to$0.98 ms).
Thus, we argue that NeuS2’s fast speed stems from a carefully integrated stack of CUDA C++ kernel implementations of ray marching, occupancy skipping, and memory layout optimization, rather than the use of TCNN.
% closed-form approximation, fused ray marching, occupancy skipping, and memory/layout optimizations), not solely from the math. 
We defer translating our method to CUDA C++ to future work.

Instead, we found that combining TCNN with a batched central finite difference (FD) approximation provides a significant computation speed improvement with minimal modifications to our PyTorch implementation. Inspired by~\cite{supernormal}, we compute surface normals (gradient) and the diagonal Hessian via batched central finite differences (FD): For each point $x\in\mathbb{R}^3$, we evaluate the SDF at six offsets $\{x \pm \varepsilon e_i\}_{i=1}^3$ in a single parallelized TCNN forward step by packing the offsets along the channel dimension, where $e_i$ are unit vectors along each axes, and $\varepsilon$ is a small offset. This ensures that only first-order (no second-order) derivatives are taken during backpropagation. As Table \ref{table:speed} shows, using the batch FD approximation leads to a significant speed-up, from 19.95\,ms to 6.02\,ms, which is $3.31\times$ faster. We will open-source all of our implementation. % of PyTorch integration of TCNN and the batched FD approximation, which replaces costly higher-order PyTorch autograd steps without a full CUDA C++ reimplementation.

%the GH module is our major win, which is the dominant speed bottleneck in our PyTorch stack (19.95\,ms); with the packed-FD implementation it drops to 6.02\,ms ($3.31`{\times}$ faster). We therefore present a \textbf{PyTorch integration of NeuS2’s MLP/backprop} (closed-form enabled and verified) together with a practical FD route that removes higher-order PyTorch autograd to fully speedup while avoiding a purely CUDA C++ implementation.
%We therefore present a \textbf{PyTorch integration of NeuS2’s MLP/backprop} (closed-form enabled and verified) together with a practical FD route that removes higher-order PyTorch autograd to fully speedup while avoiding a purely CUDA C++ implementation.

%\BL{Can we: 1) order the items in some way; and 2) rename the items from GH'' => GH-FD or something like this?}
\begin{table}[t]
\centering
\caption{\textbf{Average computation time of each module (ms)} on a single 3090Ti GPU over 1k iterations. GH(P)/GH(N)/GH(FD): Gradient/Hessian with \textbf{P}ytorch vs. \textbf{N}eus2 TCNN VS \textbf{F}inite difference (proposed). SDF/SDF(N): Pytorch vs. \textbf{N}eus2 TCNN. RGB/RGB(N): \textbf{P}ytorch VS \textbf{N}eus2 TCNN. RS: ray sampling. GS: GS aggregator, KNN, \& sparse fusion. AP: anchor points sampling.}\label{tab:speed}
\setlength{\tabcolsep}{3pt} % tighten columns (default ~6pt)
\resizebox{\columnwidth}{!}{%
\begin{tabular}{lll|ll|ll|lll}
GH(P) & GH(N) & GH(FD) & SDF  & SDF(N) & RGB  & RGB(N) & RS    & GS   & AP   \\ \hline
19.95 & 18.64 & 6.02   & 3.19 & 3.12 & 1.13 & 0.98 & 11.16 & 6.44 & 0.03
\end{tabular}
}
\label{table:speed}
\end{table}

\section{CONCLUSIONS}
We proposed SplatSDF, a novel SDf-NeRF model with an architecture-level sparse 3DGS fusion to take advantage of the 3DGS attributes and accurately manipulate SDF embeddings on surface query points. Our SplatSDF significantly improves both the accuracy and efficiency of the SDF-NeRF on geometric and photometric accuracy, and the convergence speed. 
Future work lies in real-time training of SDF-NeRF with frozen 3DGS aggregator and online updates to 3DGS.
%It outperforms all SOTA SDF-NeRFs on both geometric and photometric accuracy on two major datasets, and achieves $>3$ times convergence speed compared to the baseline, Neuralangelo.
\section*{Acknowledgement}
This work was supported by the Ministry of Trade, Industry and Energy (MOTIE), Korea, under the Strategic Technology Development Program, supervised by the Korea Institute for Advancement of Technology (KIAT) [Grant No. P0026052].

\bibliographystyle{IEEEtran}
\bibliography{main}

\clearpage
\setcounter{page}{1}
%\maketitlesupplementary
\appendix

\iffalse
\section{Project Website}
We understand that attaching an external project website violates CVPR policies. Hence, we attach the project website internally, where reviewers can easily launch the website locally with HTML. We show visualizations of the detailed 3D mesh and video comparisons on the website where the reviewers can zoom-in and rotate to check the details. Here is an instruction on how to locally launch the website.
\begin{enumerate}
    \item Unzip `website.zip'.
    \item On Visual Studio Code, install the `Live Server' extension.
    \item Under the unzipped `website' directory on Visual Studio Code, right click on `index.html', and click on `Open with Live Server'.
\end{enumerate}
\fi

\section{Relations to ``PointNeRF''}
One potential alternative to our method is to use PointNeRF \cite{pointnerf}, which takes point clouds to boost NeRF. Although the two works share the high-level idea of ``fusing prior 3D primitives to boost the target model", PointNeRF fails on our task. There are major differences persisting in the roots of the two methods. In this section, we summarize the reasons why our SplatSDF is better than PointNeRF. 

\begin{itemize}
\item \textbf{Better fusion strategies: ``Dense VS Surface''.} PointNeRF's fusion strategy does not work for SDF-NeRF. PointNeRF proposes ``Dense Point Fusion", which considers valid neighbor points for all query points on the rays. However, as shown in Figure \ref{fig:dense_vs_surface}, even though ``Dense Fusion'' might work for RGB synthesis in PointNeRF, it does not work for SDF estimation. Thus, we propose our novel ``Surface Fusion'' only at the anchor point. Our fusion strategy gets rid of the bumpy and noisy artifacts on surfaces, and is far more efficient than densely fusing all query points as PointNeRF - The efficiency improvement is further illustrated in Section \ref{section: Sparse KNN}.
%\BL{Rephrase this as `better' fusion strategy - e.g. more efficient and nothing to train!} 
\item \textbf{Better purposes: Solely RGB VS RGB+SDF.} PointNeRF is a traditional NeRF that aims at RGB synthesis, while our SplatSDF is Neural Implicit SDF for continuous SDF estimation. In other words, PointNeRF can only synthesize RGB images, our SplatSDF can achieve both continous SDF estimation and RGB synthesis. %\BL{As above. PointNeRF can only do RGB, but we can do SDF and also RGB}
\item \textbf{Better priors: 3D points VS 3DGS.} We use 3DGS as ``fusing priors'' with more attributes (opacities and shapes of 3D Gaussian primitives) than 3D points. The 3DGS opacity shares confidence priors on how likely the densities are fused to the anchor point - The higher density it carries, the more likely it is at surface. The 3DGS shape shares explicit information on the surface normals and scope of influence of Gaussian primitives, whereas point clouds (Gaussian centers) only occupies discrete and sparse coordinates without further information on shape and surface. As shown in figure \ref{fig:ablation} and table \ref{table:ablation}, leveraging these additional attributes of 3DGS enables a faster and better convergence for geometry than only using Gaussian primitive centers (point cloud). %\BL{Somewhat similar to second point}
\item \textbf{Better fusing functions: Inverse distance weighting VS 3D Gaussian weighting.} The use of 3DGS over point clouds allows the novel design of our fusing functions and 3DGS aggregator. The weight term in PointNeRF's fusing function is a basic inverse distance widely used in scattered data interpolation, while we leverage a 3D Gaussian distribution for the weight term in eq. \eqref{eq:fusion} inspired by the 2D Gaussian weights of 3DGS. We also take GS opacity into our novel fusing function as shown in eq. \eqref{eq:fusion}, as opposed to PointNeRF's confidence score obtained from the pretrained MVSNet \cite{mvsnet}.  %\BL{Similar to above - can we say that alpha blending is better than inverse distance? } 
\end{itemize}

\section{Sparse KNN Details}
\label{section: Sparse KNN}
Our goal is to take advantage of nearby GS embeddings for the SDF-MLP regression at query points. Since densely searching nearby GS for all query points leads to an unacceptable complexity, we adopt a sparse KNN from PointNeRF \cite{pointnerf}. For each view, we cast $M$ rays and sample $N$ query points along each ray. Supposing the number of GS is $\mathcal{|G|}$, a brute-force standard KNN on all query points would lead to a complexity of $O(MN|\mathcal{G}|)$. We decrease the complexity by hashing and voxelizing the 3D space into $L\textsuperscript{3}$ unit cells and reduce the complexity to $O(\frac{MN|G|}{L^3})$. Specifically, we first hash all query points and GS to unit cells, which leads to $O({|\mathcal{G}|+MN})$ complexity. Since the average number of GS in each unit cell is $\frac{|\mathcal{G}|}{L^3}$ and each query point only searches for neighbor GS within the same cell, the complexity of distance computations is $O(\frac{MN|\mathcal{G}|}{L^3})$. The overall complexity is:

\begin{equation}
    O(MN+|\mathcal{G}|) + O(\frac{MN|\mathcal{G}|}{L^3}logK) \approx O(\frac{MN|\mathcal{G}|}{L^3})
\end{equation}
where $K$ is the maximum number of nearest neighbors being considered, $\log K$ is the sorting complexity after the distance computation, but can be canceled since $K$ is a much smaller constant than the other parameters. The hashing pre-processing is canceled because $L^3\ll MN|\mathcal{G}|$. Consequently, the sparse KNN significantly reduces the complexity from $O(MN|G|)$ to $O(\frac{MN|\mathcal{G}|}{L^3})$, especially when separating fine unit cells with a high resolution $L$. In practice, we also extend the search to neighboring cells of the center cell that the query point lies in, but we ignore the complexity since the number of neighbors are commonly set to a very small number.

However, the $O(\frac{MN|\mathcal{G}|}{L^3})$ complexity is still inefficient to train an SDF-NeRF which has an extra SDF-MLP compared to the standard NeRF. As discussed in the main paper, we further reduce the complexity to $O(\frac{M|\mathcal{G}|}{L^3})$ with our proposed novel ``Surface 3DGS Fusion'' by only fusing on one query point per ray at the anchor point.

\iffalse
\section{Loss Details}
As in Section 3.3, to show the effectiveness of our architecture-level fusion, we do not seek fancy losses from geometric priors such as GS surface normals and depths as latest GS-based surface reconstruction works or the concurrent GS-SDF works, but to only use the basic ones same as Neuralangelo \cite{neuralangelo}. The losses we use are: 
\noindent
The photometric consistency loss: 
\begin{equation}
\mathcal{L}_{\text{RGB}} = \left\| \hat{\mathbf{C}} - \mathbf{C} \right\|_1
\end{equation}
where $\hat{\mathbf{C}}$ and $\mathbf{C}$ are the rendered image by SDF-NeRF and the input image, respectively. The eikonal loss on SDF estimation at query points using the fact that the gradient of SDF should be equal to 1:
\begin{equation}
\mathcal{L}_{\text{eik}} = \frac{1}{N} \sum_{i=1}^{N} \left( \left\| \nabla f_{sdf}(\mathbf{x}_i) \right\|_2 - 1 \right)^2
\end{equation}

\noindent
A curvature loss as a regression term to smooth gradient of SDF:
\begin{equation}
\mathcal{L}_{\text{curv}} = \frac{1}{N} \sum_{i=1}^{N} \left| \nabla^2 f_{sdf}(\mathbf{x}_i) \right|
\end{equation}
so the weighted sum as the final loss is:
\begin{equation}
\mathcal{L} = \mathcal{L}_{\text{RGB}} + w_{\text{eik}} \mathcal{L}_{\text{eik}} + w_{\text{curv}} \mathcal{L}_{\text{curv}}
\end{equation}
\fi

\section{Ablation Study results on ``Lego''}
We show the detailed Chamfer Distance values in $mm$ for Figure \ref{fig:ablation} in Table \ref{table:ablation}. 

\begin{table}[]
\resizebox{\columnwidth}{!}{%
\begin{tabular}{ccccccc}
Epochs                    & \begin{tabular}[c]{@{}c@{}}Neural\\ Angelo\end{tabular} & \begin{tabular}[c]{@{}c@{}}5pt\\ pcdrender\\ gsfusion\end{tabular} & \begin{tabular}[c]{@{}c@{}}1pt\\ pcdrender\\ gsfusion\end{tabular} & \begin{tabular}[c]{@{}c@{}}5pt\\ gsrender\\ gsfusion\end{tabular} & \begin{tabular}[c]{@{}c@{}}1pt\\ gsrender\\ pcdfusion\end{tabular} & \begin{tabular}[c]{@{}c@{}}1pt\\ gsrender\\ gsfusion\end{tabular} \\ \hline
\multicolumn{1}{c|}{100}  & 3.11  & 2.59  & 2.92 & 2.69 & 2.81 & \textbf{2.81}  \\
\multicolumn{1}{c|}{200}  & 2.91  & 2.60  & 3.01 & 2.50 & 2.21 & \textbf{2.54}  \\
\multicolumn{1}{c|}{300}  & 2.63  & 2.71  & 2.31 & 2.27 & 2.60 & \textbf{2.84}  \\
\multicolumn{1}{c|}{400}  & 2.62  & 2.47  & 2.37 & 2.17 & 2.08 & \textbf{2.28}  \\
\multicolumn{1}{c|}{500}  & 2.24  & 2.23  & 2.12 & 1.92 & 2.00 & \textbf{2.14}  \\
\multicolumn{1}{c|}{600}  & 2.15  & 2.08  & 1.83 & 1.84 & 1.89 & \textbf{1.79}  \\
\multicolumn{1}{c|}{700}  & 1.97  & 1.94  & 1.73 & 1.68 & 1.74 & \textbf{1.71}  \\
\multicolumn{1}{c|}{800}  & 1.96  & 1.93  & 1.84 & 1.68 & 1.76 & \textbf{1.61}  \\
\multicolumn{1}{c|}{900}  & 1.88  & 1.69  & 1.72 & 1.54 & 1.62 & \textbf{1.48}  \\
\multicolumn{1}{c|}{1000} & 1.77  & 1.81  & 1.70 & 1.49 & 1.57 & \textbf{1.41}  \\
\multicolumn{1}{c|}{1100} & 1.76  & 1.72  & 1.56 & 1.54 & 1.52 & \textbf{1.42}  \\
\multicolumn{1}{c|}{1200} & 1.73  & 1.70  & 1.54 & 1.45 & 1.47 & \textbf{1.39}  \\
\multicolumn{1}{c|}{1500} & 1.71  & 1.62  & 1.54 & 1.42 & 1.42 & \textbf{1.35}  \\
\multicolumn{1}{c|}{2000} & 1.62  & 1.56  & 1.43 & 1.38 & 1.34 & \textbf{1.27}  \\
\multicolumn{1}{c|}{2500} & 1.54  & 1.52  & 1.38 & 1.34 & 1.35 & \textbf{1.24}  \\
\multicolumn{1}{c|}{3000} & 1.60  & 1.49  & 1.34 & 1.34 & 1.26 & \textbf{1.19}                                                             
\end{tabular}
}
\caption{\textbf{Values of Figure \ref{fig:ablation}, } Comparison on geometric accuracy (Chamfer Distance in mm) at different training epochs on “Lego”.}
\label{table:ablation}
\end{table}

\section{Results on NeRF Sythetic Dataset}

We show the qualitative and quantitative comparison of ``Lego'' in the main paper. In this section, we show the qualitative and quantitative comparison on another difficult scene from the NeRF Synthetic Dataset, ``Ship''.

\begin{figure}[t]
    \centering
    \includegraphics[width=\columnwidth]{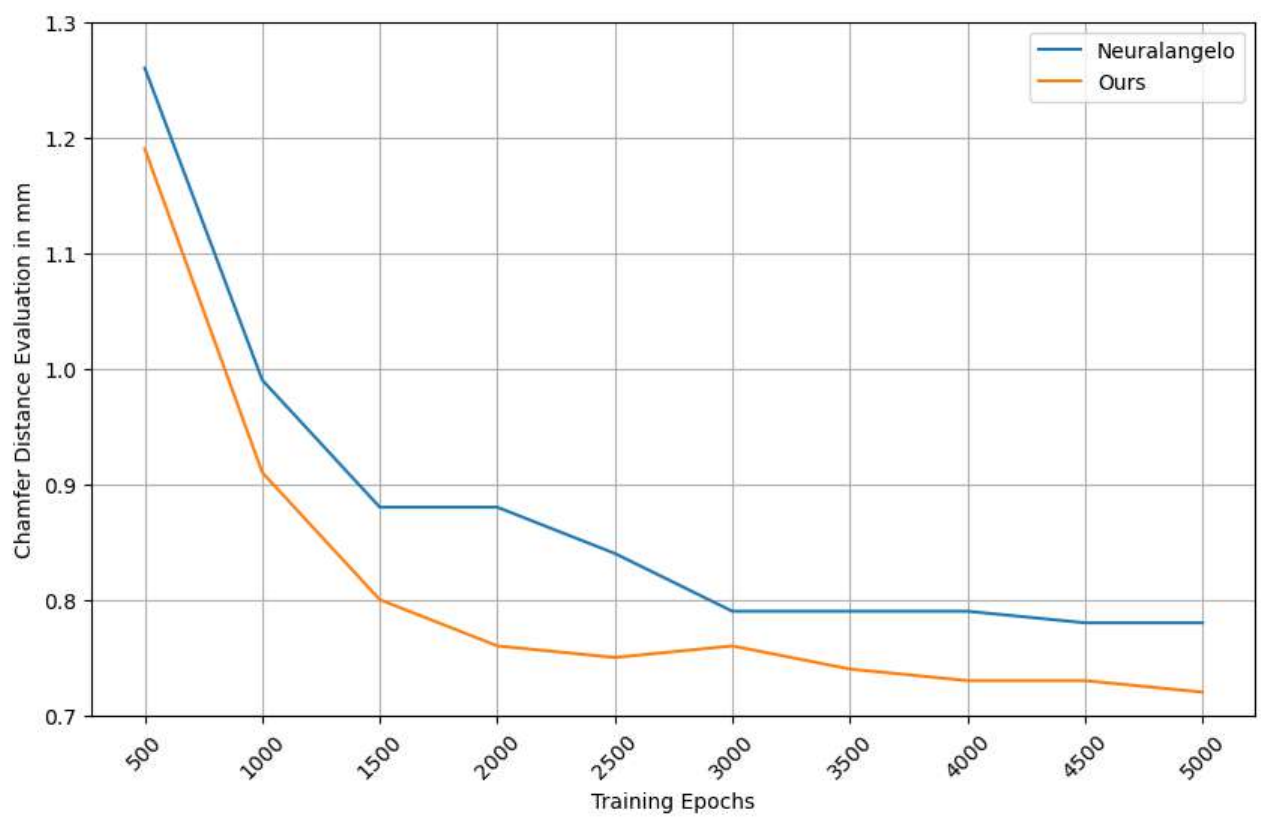}
    \caption{\textbf{Comparison of geometric accuracy on ``Ship''}. Our method achieves faster convergence to better accuracy than the baseline ``Neuralangelo''. Exact values are summarized in the table .}
    \label{fig:ablation_ship}
\end{figure}

%Figure \ref{fig:ship} shows visual comparisons of our SplatSDF to the best baseline Neuralangelo at different training epochs. 
Figure \ref{fig:ablation_ship} and Table \ref{tab:ablation_ship} show the numerical comparison in Chamfer Distance. Both the qualitative and quantitative results prove our faster and better convergence to complex shapes.

\begin{table}[]
\begin{tabular}{ccc}
\multicolumn{1}{l}{Epochs} & \multicolumn{1}{l}{Neuralangelo} & \multicolumn{1}{l}{SplatSDF (Ours)} \\ \hline
\multicolumn{1}{c|}{500}   & 1.26                             & \textbf{1.19}                                \\
\multicolumn{1}{c|}{1000}  & 0.99                             & \textbf{0.91}                                \\
\multicolumn{1}{c|}{1500}  & 0.88                             & \textbf{0.80}                                \\
\multicolumn{1}{c|}{2000}  & 0.88                             & \textbf{0.76}                                \\
\multicolumn{1}{c|}{2500}  & 0.84                             & \textbf{0.75}                                \\
\multicolumn{1}{c|}{3000}  & 0.79                             & \textbf{0.76}                               \\
\multicolumn{1}{c|}{3500}  & 0.79                             & \textbf{0.74}                               \\
\multicolumn{1}{c|}{4000}  & 0.79                             & \textbf{0.73}                               \\
\multicolumn{1}{c|}{4500}  & 0.78                             & \textbf{0.73}                               \\
\multicolumn{1}{c|}{5000}  & 0.78                             & \textbf{0.72}                              
\end{tabular}
\caption{\textbf{Values of Figure \ref{fig:ablation_ship}, } Comparison on geometric accuracy (Chamfer Distance in mm) at different training epochs on “Ship”.}
\label{tab:ablation_ship}
\end{table}

\section{Results on DTU Dataset}

\begin{table*}[]
\resizebox{\textwidth}{!}{
\begin{tabular}{llllllllllllll}
Scan ID   & 24    & 37    & 40    & 55    & 63    & 65    & 69    & 83    & 105   & 106   & 110   & 114   & Mean  \\ \hline
RegSDF \cite{regsdf}    & 24.78 & 23.06 & 23.47 & 22.21 & 28.57 & 25.53 & 21.81 & 28.89 & 27.91 & 24.71 & 25.13 & 26.84 & 25.24 \\
NeuS \cite{neus}     & 26.62 & 23.64 & 26.43 & 25.59 & 30.61 & 32.83 & 29.24 & 33.71 & 31.97 & 32.18 & 28.92 & 28.41 & 29.18 \\
VolSDF \cite{volsdf}   & 26.28 & 25.61 & 26.55 & 26.76 & 31.57 & 31.50 & 29.38 & 33.23 & 32.13 & 33.16 & 31.49 & 30.33 & 29.83 \\
NeRF \cite{nerf}     & 26.24 & 25.74 & 26.79 & 27.57 & 31.96 & 31.50 & 29.58 & 32.78 & 32.08 & 33.49 & 31.54 & 31.00 & 30.02 \\
Neuralangelo \cite{neuralangelo} & 30.64 & 27.78 & 32.70 & 34.18 & 35.15 & 35.89 & 31.47 & 36.82 & 35.92 & 36.61 & 32.60 & 31.20 & 33.41 \\ \hline
\begin{tabular}[c]{@{}l@{}}Neuralangelo \cite{neuralangelo} (we trained)\end{tabular} & 29.65 & 30.45 & 28.90 & 28.42 & 30.24 & 28.36 & 28.15 & 29.65 & 28.66 & 28.57 & 29.00 & 28.38 & 29.04 \\
Ours     & \textbf{29.63} & \textbf{30.39} & \textbf{29.75} & \textbf{28.44} & \textbf{31.24} & \textbf{28.74} & \textbf{29.00} & \textbf{29.61} & \textbf{29.41} & \textbf{28.83} & \textbf{29.45} & \textbf{28.68} & \textbf{29.43}
\end{tabular}
}
\caption{\textbf{Photometric evaluation on DTU dataset in PSNR.} The results of the first 5 methods (RegSDF, NeuS, VolSDF, NeRF and Neuralangleo) are directly from Neuralangelo's paper. Neuralangelo did not release their trained checkpoints, we retrain by ourself but cannot reproduce their paper's results using their code. The bottom two rows show the results of Neuralangelo trained by ourself, and our SplatSDF. }
\label{tab:dtu_psnr}
\end{table*}

In the main paper, we show the geometric evaluation (in Chamfer Distance) on the DTU Dataset. In this section, we show the visual comparison in Figure \ref{fig:dtu_vis} and photometric evaluations in Table \ref{tab:dtu_psnr}. We train Neuralangelo by ourself since there are no checkpoints released by Neuralangelo. Both Neuralangelo and our SplatSDF are trained to fully converge. For our SplatSDF, we use the same checkpoint as the evaluation in Table \ref{table:dtu_cd}. Since visual differences are difficult to observe in most of the cases, we select some representative details as shown in Figure \ref{fig:dtu_vis}. Neuralangelo fails to reconstruct some difficult details, while our SplatSDF can reconstruct some difficult details at a high quality. Table \ref{tab:dtu_psnr} shows the geometric evaluation in PSNR on the DTU Dataset. We cannot reproduce the same results that Neuralangelo's paper proposes, so we also show Neuralangelo's result trained by ourself. Our SplatSDF obtains a higher PSNR than Neuralangelo on all of the scenes.

\section{Training and Inference Efficiency}
We build our SplatSDF upon Neuralangelo \cite{neuralangelo}, where training speed on a single Nvidia 3090Ti GPU is $\sim$ 7FPS with the default training settings - 512 rays per image and 128 query points per ray. When using our Surface 3DGS Fusion, the speed is slightly affected by the number of valid 3DGS being fused and valid rays. The normalized unit sphere's radius is 1. We empirically set the voxel size of sparse KNN to be 0.005, and K is 4. The training speed of our SplatSDF is $\sim$ 5.5 FPS on a single Nvidia 3090Ti. Since we only do 3DGS Fusion for training, we do not change the efficiency of the original SDF-NeRF, and in this work, the inference speed is exactly the same as Neuralangelo.

\section{Scene Selections}
In this section, we explain more details of the scenes we choose for evaluation. For a fair comparison, since we completely build our algorithm on Neuralangelo \cite{neuralangelo}, we train Neuralangelo to make sure we reproduce comparable results that the Neuralangelo paper proposes as Neuralangelo does not release their well-trained checkpoints. Since the DTU Dataset only scans on one side of the object by swinging at a limited region, SDF-NeRF has no access to the back (or unseen) sides of the object, and it is necessary to crop Neuralangelo's results to get comparable results as their paper proposes. Please refer to the Neuralangelo official GitHub repository to understand the reason for cropping \url{https://github.com/NVlabs/neuralangelo/issues/93}. We use 12 but not all of the commonly used 15 scenes for the DTU evaluation because we cannot reproduce comparable results using their code implementation on scene 97, 118 and 122 that Neuralangelo proposes. 

\section{Resources of Comparison}
For the comparison results in Table \ref{table:dtu_cd} and Table \ref{table:nerfsyn_cd}, we use the results from the latest paper or the original paper directly. In Table \ref{table:dtu_cd}, for SDF-NeRF methods, we get the results of NeRF \cite{nerf}, VolSDF \cite{volsdf}, NeuS \cite{neus}, HF-NeuS \cite{hfneus}, RegSDF \cite{regsdf}, NeuralWarp \cite{neuralwarp} and Neuralangelo \cite{neuralangelo} from Neuralangelo. We get the results of COLMAP \cite{colmap}, MVSDF\cite{mvsdf}, NeuS-12 \cite{neus} from TUVR \cite{tuvr}. Furthermore, the results for TUVR \cite{tuvr} and DebSDF are from \cite{debsdf2024}. The results UNISURF \cite{unisurf} and Geo-NeuS \cite{geoneus} are from Geo-NeuS. The results of OAV \cite{oav} and PET-NeuS \cite{petneus} are from their own papers. For GS-based Surface Reconstruction methods, we get Scalffold-GS \cite{scaffoldgs} from GSDF \cite{gsdf}. We get 3DGS \cite{3DGS}, SuGAR \cite{sugar}, \cite{gaussiansurfels}, 2DGS \cite{2DGS} and Gaussian Opacity Field \cite{gaussianopaictyfield} from Gaussian Opacity Field. For the two concurrent GS-based Surface Reconstruction 3DGSR \cite{3dgsr} and GSDF \cite{gsdf}, we get the results from their original paper. In Table \ref{table:nerfsyn_cd}, since Neuralangelo does not evaluate on NeRF Synthetic dataset, we implement Neuralangelo and our SplatSDF by ourself. For the Chamfer Distance in the top part, we get the results of VolSDF \cite{volsdf}, NeuS \cite{neus}, NeRO \cite{nero}, BakedSDF \cite{bakedsdf}, NeRF2Mesh \cite{nerf2mesh}, RelightableG \cite{relightablegaussian}, and 3DGSR \cite{3dgsr} from 3DGSR. We get the result of 
HF-NeuS \cite{hfneus} from its paper. For the PSNR 
 at the bottom part, we get the results of NeRF \cite{nerf}, VolSDF \cite{volsdf}, NeuS \cite{neus}, HF-NeuS \cite{hfneus}, PET-NeuS \cite{petneus} from PET-NeuS. We get the results of NeRO \cite{nero}, BakedSDF \cite{bakedsdf}, NeRF2Mesh \cite{nerf2mesh}, Mip-NeRF \cite{mipnerf}, 3DGS \cite{3dgsr} and Instant-NGP \cite{instantngp} from 3DGSR \cite{3dgsr}.

\end{document}